\documentclass[11pt]{article}

\usepackage[preprint]{acl}

\usepackage{times}
\usepackage{latexsym}

\usepackage[T1]{fontenc}

\usepackage[utf8]{inputenc}

\usepackage{microtype}

\usepackage{inconsolata}

\usepackage{graphicx}

\usepackage{booktabs}
\usepackage{multirow}
\usepackage{CJKutf8}
\usepackage[utf8]{inputenc} 
\usepackage{url}            
\usepackage{booktabs}       
 \usepackage{amsmath} 
\usepackage{graphicx} 
\usepackage{float}    

\usepackage{amsmath}
\usepackage{comment}
\usepackage{tikz}
\usetikzlibrary{arrows.meta,calc}
\usepackage{tcolorbox}
\usepackage{fontawesome5}
\usepackage{pifont}
\newcommand{\cmark}{\textcolor{green!45!black}{\ding{51}}}
\newcommand{\xmark}{\textcolor{red!65!black}{\ding{55}}}
\tcbuselibrary{listings}
\usepackage{color}  

\newcommand{\legalricon}{%
\raisebox{-0.18em}{\includegraphics[height=1.35em]{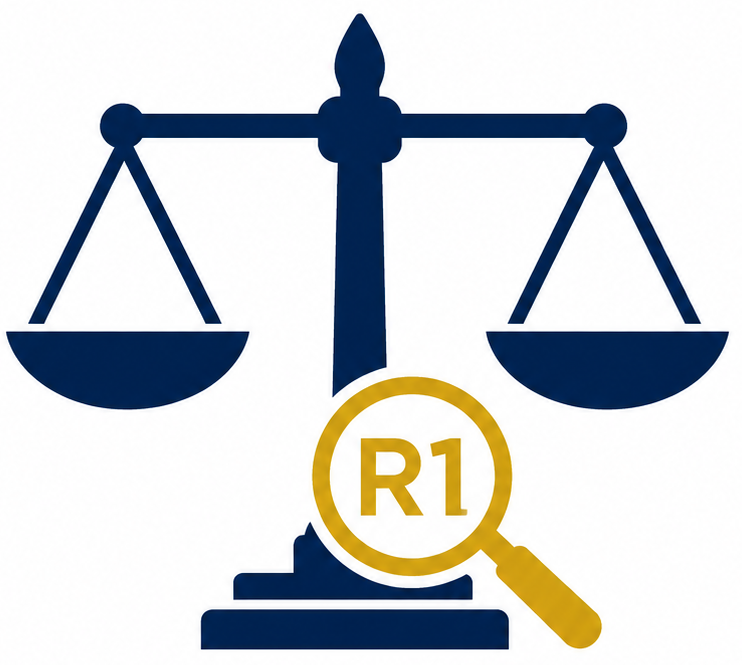}}}
\title{\legalricon\ Legal-R1: Agentic Retrieval for Evidence-Based Legal Reasoning}

\author{
 \textbf{Huaiwen Pang\textsuperscript{1,*}},
 \textbf{Jiale Han\textsuperscript{2,*,$\dagger$}},
 \textbf{Junying Chen\textsuperscript{1}},
 \textbf{Wenjun Xiao\textsuperscript{3}},
 \textbf{Benyou Wang\textsuperscript{1,2,$\dagger$}}
 \\
 \textsuperscript{1}The Chinese University of Hong Kong, Shenzhen,
 \\
 \textsuperscript{2}Shenzhen Loop Area Institute, 
 \textsuperscript{3}Shenzhen Shenten Digital Technology Co., Ltd.
 \\
 \texttt{huaiwenpang@link.cuhk.edu.cn},
 \texttt{jialehan@slai.edu.cn},
 \texttt{junying.chen.cs@gmail.com},\\
 \texttt{xiaowenjun@fadaifu.net},
 \texttt{wangbenyou@cuhk.edu.cn}
 }

\begin{document}
\maketitle
\begingroup\renewcommand{\thefootnote}{\fnsymbol{footnote}}\footnotetext[1]{Equal contribution.}\footnotetext[2]{Corresponding authors.}\endgroup

\begin{abstract}

Legal question answering requires answers that are not only plausible, but also grounded in legally valid and citable evidence. Standard retrieval-augmented generation often relies on one-shot semantic matching, which is insufficient when legal evidence must be selected according to authority, applicability, and diversity across heterogeneous sources. We introduce \textbf{Legal-R1}, an agentic retrieval framework for evidence-based legal reasoning. Legal-R1 decomposes legal reasoning into a multi-agent workflow, where a Lawyer agent learns when to retrieve, which legal source to query, what evidence to seek, and when the collected evidence is sufficient. A Legal Assistant agent retrieves from normative legal sources and precedent cases with authority-aware and diversity-aware selection, while a Judge agent generates the final answer from the complete evidence trajectory. We post-train only the Lawyer agent with masked supervised fine-tuning and trajectory-aware reinforcement learning. Experiments on multiple Chinese legal benchmarks show that Legal-R1 improves over general-purpose, legal-specific, and retrieval-augmented baselines.\footnote{The code, model, and data are available at \url{https://github.com/killerqueen-wonder/Legal-R1}.}
\end{abstract}

\section{Introduction}
\begin{figure}[t]
    \centering
    \includegraphics[width=0.9\columnwidth]{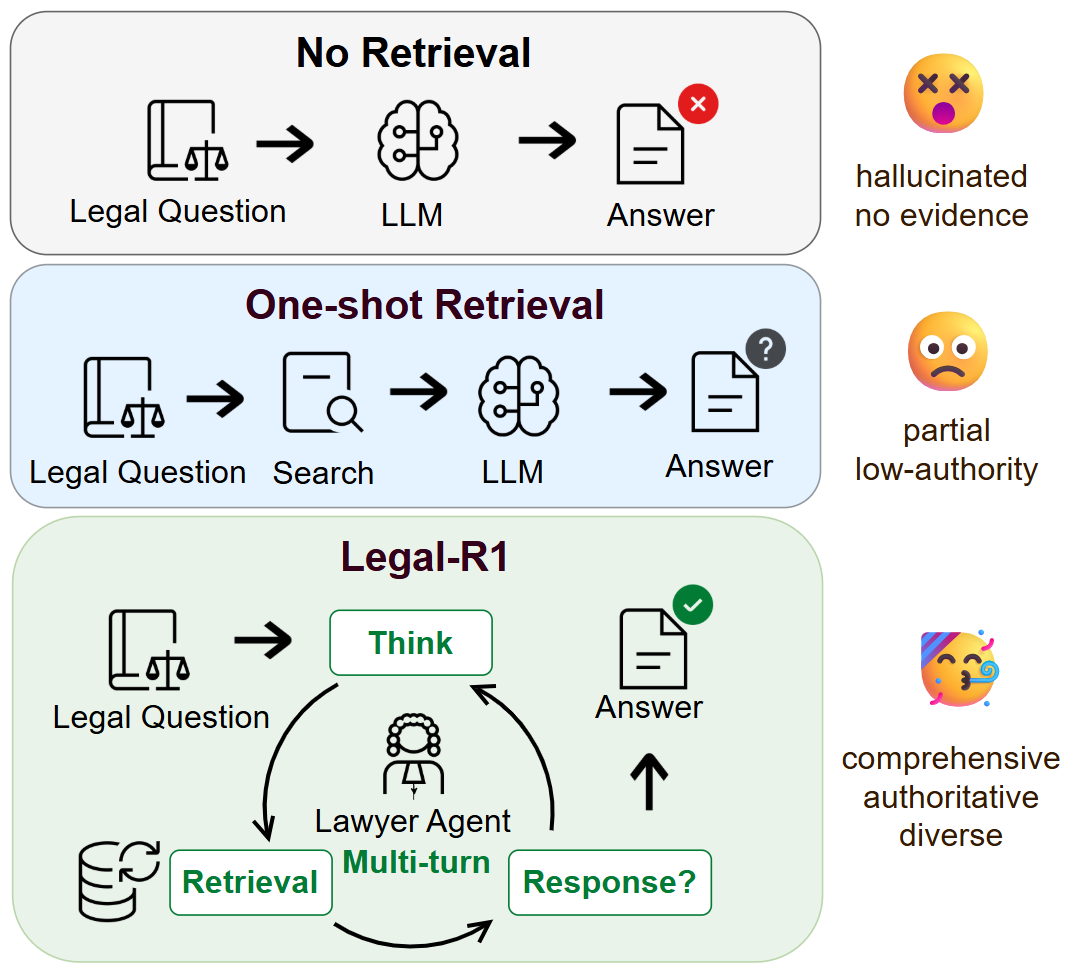}
    \caption{Motivation of Legal-R1. Legal-R1 replaces one-turn semantic matching with multi-turn evidence seeking over norms and cases.}
    \label{fig:intro_motivation}
\end{figure}
Legal question answering requires more than fluent generation. A trustworthy legal answer must be grounded in legal evidence that is valid, applicable, and citable \citep{wiratunga2024cbr,louis2024interpretable,xiao2026structuring}. This evidence is often distributed across heterogeneous sources, including laws, judicial interpretations, and precedent cases. It must also satisfy domain-specific constraints such as authority hierarchy, temporal validity, jurisdictional fit, and applicability to the target facts \citep{van2017concept}. These requirements make legal question answering a problem of evidence construction rather than plain information lookup.

Existing large language models still struggle with this setting. Parametric legal knowledge can be incomplete, outdated, or unsupported by explicit citations \citep{dahl2024large,van2017concept}. Retrieval-augmented generation improves grounding by retrieving external documents \citep{lewis2020retrieval}, but many conventional pipelines rely on a fixed set of retrieved passages without adaptively deciding whether further retrieval is needed \citep{asai2024self}. This is not sufficient for legal reasoning. A semantically similar case may still fail to match the relevant legal issue \citep{li2024lecardv2}. Other retrieved cases may come from lower-authority courts or duplicate the same sentencing outcome. Legal retrieval therefore cannot be reduced to finding similar text. It must identify legally valid evidence that can justify the answer. Figure~\ref{fig:intro_motivation} illustrates this distinction. General agentic retrieval asks what information is needed to answer a question. Legal agentic retrieval asks what legal norms and precedents are needed to support a verifiable answer. The agent must decide whether retrieval is needed, which legal source to query, what evidence to retrieve, and whether the collected evidence is sufficient.

We propose \textbf{Legal-R1}, an agentic retrieval framework for evidence-based legal reasoning. Legal-R1 decomposes legal reasoning into three roles. The \textit{Lawyer agent} plans multi-turn evidence seeking. It identifies missing legal evidence, chooses a retrieval action, formulates source-specific queries, and decides when to stop. The \textit{Legal Assistant agent} executes retrieval over heterogeneous legal knowledge bases. It searches legal norms, including laws and judicial interpretations, and retrieves precedent cases with authority-aware and diversity-aware selection. The \textit{Judge agent} generates the final answer from the legal question and the complete retrieval trajectory. This workflow makes evidence collection explicit and encourages answers to be grounded in comprehensive, authoritative, and diverse legal evidence.

We train only the Lawyer agent. This design focuses learning on the core difficulty of legal agentic retrieval, namely evidence-seeking decisions. It also decouples retrieval planning from answer generation, so the trained Lawyer agent can be paired with different Judge agents in downstream applications. To train the Lawyer agent, we synthesize multi-turn retrieval trajectories from a structured legal QA dataset. We first use masked supervised fine-tuning to teach the retrieval protocol while preventing the model from memorizing retrieved evidence. We then apply trajectory-aware reinforcement learning to improve source selection, query construction, and stopping behavior.

Experiments on Chinese legal benchmarks validate the effectiveness of Legal-R1. Compared with existing RAG-based legal systems, Legal-R1 achieves stronger performance on LexEval and LawBench, showing that actively planning retrieval is more effective than attaching a fixed retriever to an LLM. Ablation and retrieval-quality analyses further show that post-training improves evidence-seeking behavior and that authority-aware and diversity-aware precedent reranking improves the reliability of retrieved evidence. These results support our central claim that legal reasoning benefits from learning an explicit evidence-seeking policy.

Our contributions are summarized as follows:
\begin{itemize}
    \item We introduce a multi-agent workflow for evidence-based legal reasoning, where the Lawyer, Legal Assistant, and Judge agents separately handle evidence planning, legal retrieval, and answer generation. This workflow supports comprehensive, authoritative, and diverse evidence grounding.
    \item We propose Legal-R1, an agentic legal retrieval framework that post-trains the Lawyer agent to learn an autonomous evidence-seeking policy, including whether to retrieve, which source to query, what evidence to retrieve, and when to stop.
    \item We conduct experiments and ablations on multiple Chinese legal benchmarks, demonstrating that Legal-R1 improves legal reasoning over retrieval-augmented baselines and that its gains come from both learned retrieval policies and legal-evidence-aware reranking.
\end{itemize}

\section{Related Work}

\paragraph{Retrieval-Augmented Generation.}
Retrieval-Augmented Generation (RAG) mitigates factual hallucinations by dynamically fetching external knowledge, evolving from basic embedding lookups to hybrid frameworks that merge lexical and semantic search \cite{lewis2020retrieval,karpukhin2020dense}. To handle massive document pools, recent paradigms employ large language models as re-rankers to prioritize relevant passages \cite{sun2023chatgpt,nogueira2020document}. However, conventional RAG systems treat documents as independent text blocks, relying on surface-level semantic overlaps while completely ignoring domain-specific structured meta-attributes like judicial hierarchies or sentencing variations \cite{gao2023retrieval}. In contrast, we propose a specialized judicial re-ranking mechanism integrating an authority hierarchy score and a punishment severity index (PSI). By incorporating these domain priors into a modified maximal marginal relevance formulation \cite{carbonell1998use}, our approach simultaneously optimizes both statutory authority and retrieval diversity.

\begin{figure*}[t!] 
    \centering 
    \includegraphics[width=0.8\textwidth]{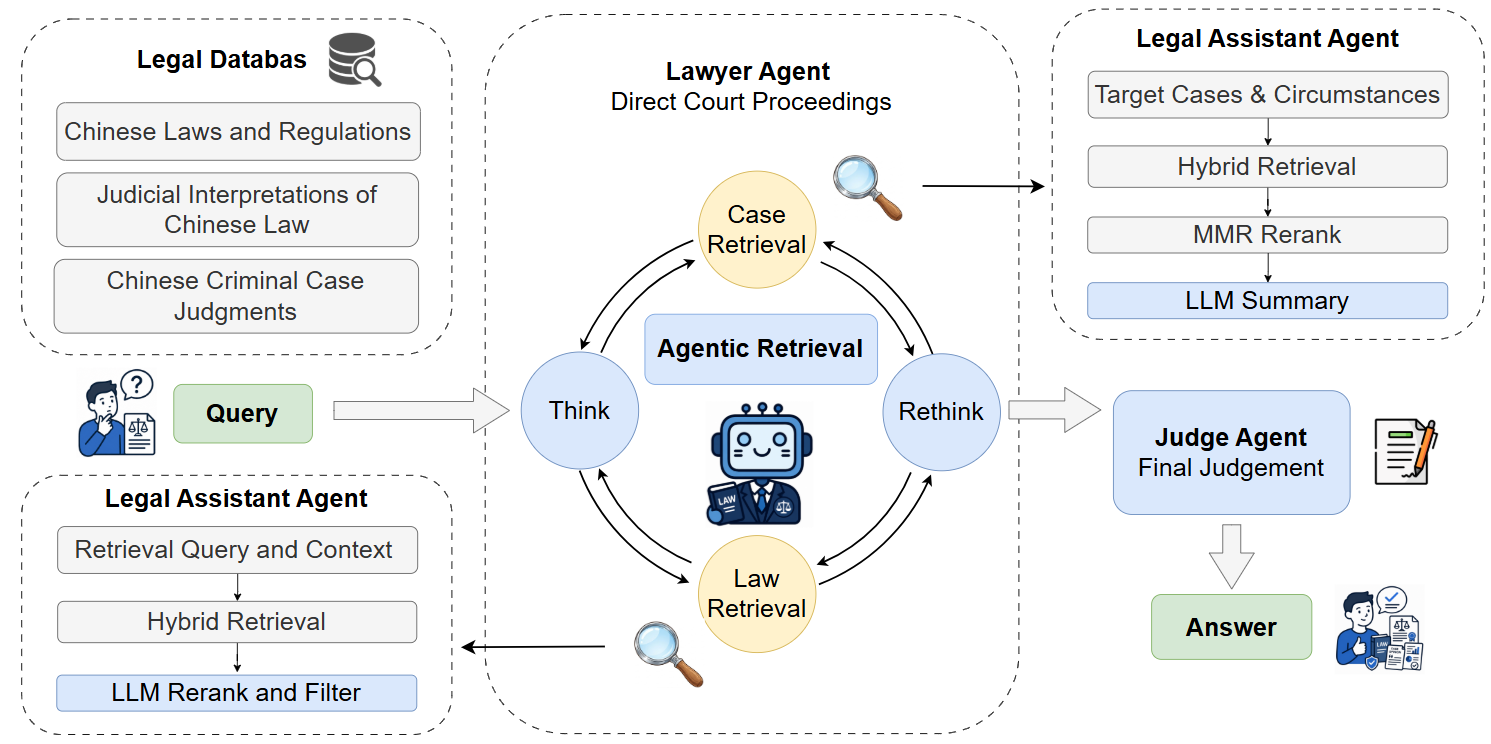} 
    \caption{Overview of the proposed agentic legal retrieval workflow.} 
    \label{fig:agentic_workflow} 
\end{figure*}

\paragraph{Tool-Augmented Agents.}
Tool-augmented agents empower language models to autonomously execute external APIs, reason through complex environments, and solve multi-step tasks \cite{yao2022react,schick2023toolformer}. Recent breakthroughs leverage post-training reinforcement learning—such as Group Relative Policy Optimization (GRPO)—to scale test-time compute and cultivate long-term thinking behaviors for autonomous error correction \cite{shao2024deepseekmath,guo2025deepseek}. Despite their success, existing frameworks heavily rely on deterministic, mathematically verifiable environments where outcomes are definitively checked by code interpreters or symbolic solvers \cite{cobbe2021training}. When applied to non-deterministic, knowledge-dense fields like law, they suffer from severe supervision bottlenecks due to the absence of absolute verification functions \cite{hendrycks2021measuring}. Our approach bridges this gap through a hierarchical multi-agent workflow embedded in a continuous deliberation loop. We adapt GRPO to this non-deterministic tool-use paradigm via a multi-tier dense reward function anchored by a fine-tuned critique model \cite{ouyang2022training}.

\paragraph{LLMs and Agents in the Legal Domain.}
The deployment of large language models in computational law has transformed legal natural language processing \cite{xiao2018cail2018}. Existing efforts either prompt general models with legal instructions or fine-tune them on domain-specific corpora \cite{fei2024lawbench}. Recently, multi-agent legal pipelines have emerged to simulate role-differentiated behaviors, alongside synthetic data pipelines to compile large-scale training trajectories \cite{yue2023disc,wang2023self}. Nevertheless, standard legal models function within rigid, single-turn setups that struggle with fluid reasoning, while automated data generation remains highly susceptible to training-test contamination and option overlaps \cite{xu2024benchmark}. Our work addresses these gaps by introducing an end-to-end automated data synthesis and purification pipeline. This framework implements strict algorithmic safeguards, including fingerprint-based contamination filtering and heuristic negative sampling for distractors, ensuring data purity and legal logical uniqueness.

\section{Method}
\subsection{Problem Formulation}
Given a legal question $q$, the task is to generate an evidence-grounded answer $a$ based on a heterogeneous legal knowledge base $\mathcal{K}$. In our setting, $\mathcal{K}$ consists of legal norms and judicial precedents.
\begin{equation*}
\mathcal{K}=\mathcal{K}_{norm}\cup\mathcal{K}_{case},
\end{equation*}
where $\mathcal{K}_{norm}$ contains laws, regulations, and judicial interpretations, and $\mathcal{K}_{case}$ contains criminal judgment cases.

Legal evidence is not judged by relevance alone. A high-quality answer should be grounded in evidence that is \textit{comprehensive}, \textit{authoritative}, and \textit{diverse}. It should cover the key legal issues in $q$, rely on valid and authoritative legal sources, and include analogous precedents that reflect meaningful variation in fact patterns and judicial outcomes. The central challenge is therefore to decide what evidence should be retrieved, from which legal source, and when the collected evidence is sufficient for answering.

\subsection{Agentic Legal Retrieval}

As shown in Figure~\ref{fig:agentic_workflow}, Legal-R1 addresses evidence-based legal reasoning through three role-specialized components: a Lawyer Agent, a Legal Assistant Agent, and a Judge Agent. The Lawyer Agent is the central planner. It analyzes the user question, identifies missing legal evidence, selects a source-specific retrieval action, formulates or revises the query, and decides whether the accumulated evidence is sufficient to stop. The Legal Assistant Agent executes these requests over heterogeneous legal knowledge bases: it retrieves and reranks statutes and judicial interpretations. The Judge Agent receives the legal question and the complete evidence trajectory, and synthesizes the final evidence-grounded answer from this collected evidence. This role separation makes the learnable object explicit: Legal-R1 trains the Lawyer Agent to improve the evidence-seeking policy, while the Legal Assistant and Judge Agents provide retrieval and answer synthesis services.

\subsubsection{Lawyer Agent}
Given $q$, the Lawyer agent identifies the legal issues, determines what type of evidence is missing, and constructs a multi-turn retrieval trajectory.
\begin{equation*}
\tau = (r_1, u_1, s_1, e_1, \dots, r_T, u_T, s_T, e_T)
\end{equation*}
where $r_j$ is the reasoning state, $u_j$ is the retrieval action, $s_j$ is the search query, and $e_j$ is the retrieved evidence at turn $j$. The action space is defined as
\begin{equation*}
u_j \in \{\textsc{Statute}, \textsc{Precedent}, \textsc{Stop}\}
\end{equation*}
Here, \textsc{Statute} requests legal provisions or judicial interpretations, \textsc{Precedent} requests analogous criminal cases, and \textsc{Stop} indicates that the current evidence is sufficient for answering. The Lawyer agent therefore decides whether the next step should clarify the legal basis, compare precedents, refine a failed query, or terminate retrieval. After each returned evidence unit, it performs a rethink step to assess whether the evidence resolves the current uncertainty and whether the trajectory should continue with the same source, switch to another source, or stop.

\subsubsection{Legal Assistant Agent}
The Legal Assistant agent executes the Lawyer agent's retrieval requests over $\mathcal{K}$ and returns compact evidence units.

\noindent\textbf{Intent-Aware Statute Retrieval.}
For \textsc{Statute}, the Lawyer agent provides a search query and a short retrieval intent or context, such as locating a sentencing range, identifying the elements of an offense, or finding an exception clause. The Legal Assistant agent first retrieves candidate provisions from $\mathcal{K}_{norm}$ using the query. It then reranks the candidates with an LLM-based normative reranker $\mathcal{L}_{norm}$ conditioned on the retrieval intent, so that the returned provisions match the specific legal role requested by the Lawyer agent.

\noindent\textbf{Authority-Aware Diverse Precedent Retrieval.}
For \textsc{Precedent}, the Legal Assistant agent retrieves a candidate set $C$ from $\mathcal{K}_{case}$ using the target fact description and optional circumstances. Since semantic similarity alone may return low-authority or outcome-redundant cases, we adapt Maximal Marginal Relevance (MMR)~\citep{carbonell1998use} to select precedents with both court authority and sentencing-outcome diversity. For each candidate $d_i \in C$, the base score is
\begin{equation*}
\text{Base}(d_i)=\alpha \cdot \text{Rel}(d_i)
 +(1-\alpha)\cdot \text{Auth}(d_i)
\end{equation*}
where $\text{Rel}(d_i)$ is the retrieval relevance score, $\text{Auth}(d_i)$ is the normalized court-authority score, and $\alpha$ controls their trade-off. Given selected cases $S$, the final MMR-style score is
\begin{equation*}
M(d_i \mid S)= \text{Base}(d_i) - \beta \cdot
\max_{d_j \in S}\text{Sim}_{\text{PSI}}(d_i,d_j)
\end{equation*}
where $\beta$ controls the diversity penalty and $\text{Sim}_{\text{PSI}}(d_i,d_j)=1-\frac{|\text{PSI}_i-\text{PSI}_j|}{d_{\max}+\epsilon}$ measures similarity in sentencing severity. Here, $d_{\max}$ normalizes PSI distance and $\epsilon$ ensures numerical stability. Section~\ref{knowledge_base} details the authority annotation and PSI construction. After selection, an LLM-based case summarizer $\mathcal{L}_{case}$ condenses each precedent into a compact evidence unit with key facts, court level, judgment outcome, and sentencing information. 

\subsubsection{Judge Agent}
The Judge agent directly generates the final evidence-grounded answer. We denote the answer-generation LLM as $\mathcal{L}_{Judge}$. Given the legal question $q$ and the evidence collected along the retrieval trajectory $\tau$, the Judge agent produces
\begin{equation*}
a=\mathcal{L}_{Judge}(q,\tau)
\end{equation*}
The prompt instructs $\mathcal{L}_{Judge}$ to answer based on the retrieved legal norms and precedents, and the full prompt is provided in the appendix~\ref{sec:appendix_prompts}.

\subsection{Lawyer Agent Training}
We train the Lawyer agent to learn an active evidence-seeking policy for legal retrieval. It learns when to retrieve, what evidence to seek, how to revise source-specific queries, and when to stop. Training only the Lawyer agent decouples evidence collection from answer generation, allowing the learned policy to be paired with different Judge agents in downstream applications.

\noindent\textbf{Trajectory Synthesis.}
We synthesize tool-use trajectories by simulating the inference-time interaction between the Lawyer agent and the Legal Assistant agent. Given a legal question $q$, a teacher model $\mathcal{L}_{Teacher}$ plays the Lawyer agent and generates a reasoning state $r_j$, a retrieval action $u_j$, and a source-specific query $s_j$. When $u_j \neq \textsc{Stop}$, the Legal Assistant agent uses $s_j$ to invoke the corresponding retrieval module and returns evidence $e_j$. The teacher conditions on the updated context to decide whether to refine the query, switch the evidence source, or stop retrieval. The process terminates when $\mathcal{L}_{Teacher}$ emits \textsc{Stop} and produces the final answer $a$. Each training instance records the complete evidence-seeking process.
\begin{equation*}
\begin{aligned}
D = \{ &(q^{(i)}, r_1^{(i)}, u_1^{(i)}, s_1^{(i)}, e_1^{(i)}, \dots, \\
&r_{T_i}^{(i)}, u_{T_i}^{(i)}, s_{T_i}^{(i)}, e_{T_i}^{(i)}, a^{(i)}) \}_{i=1}^{N}
\end{aligned}
\end{equation*}
where $D$ is the synthesized trajectory dataset, $N$ is the number of training instances, and $T_i$ is the number of retrieval turns in instance $i$.

\noindent\textbf{Evidence-Masked Supervised Fine-Tuning.}
We use supervised fine-tuning to teach the Lawyer agent the protocol of agentic legal retrieval. Each sample in $D$ is serialized into a token sequence $x=(x_1,\ldots,x_L)$. We optimize the causal language modeling loss with an evidence mask:
\begin{equation*}
\mathcal{L}_{\text{SFT}}
=-\sum_{t=1}^{L} m_t \cdot \log p_{\theta}(x_t \mid x_{<t}),
\end{equation*}
where $m_t=0$ for tokens belonging to retrieved evidence $e_j$ and $m_t=1$ otherwise. This masking prevents the model from learning to reproduce retrieved statutes or cases, and instead focuses training on reasoning states, retrieval actions, and query revision.

\noindent\textbf{Trajectory-Aware Reinforcement Learning.}
Starting from the SFT model, we use Group Relative Policy Optimization (GRPO)~\citep{shao2024deepseekmath} to further train the Lawyer agent. For each question $q$, the current model samples a retrieval trajectory $\hat{\tau}$ and a final answer $\hat{a}$. The reward compares them with the synthesized trajectory $\tau$ and reference answer $a$ in $D$. It encourages the Lawyer agent to collect useful evidence, use valid queries and formats, and avoid unnecessary retrieval turns. The total reward is
\begin{equation*}
\begin{aligned}
R(\hat{\tau}, \hat{a} \mid q,\tau,a)
=& \, w_1 R_{\text{acc}}(\hat{a}, a)
+ w_2 R_{\text{align}}(\hat{\tau}, \tau) \\
&+ w_3 R_{\text{gain}}(\hat{e}, a)
+ w_4 R_{\text{query}}(\hat{s}) \\
&+ R_{\text{fmt}}(\hat{\tau}, \hat{a})
- \rho(\hat{\tau}, \hat{a}) - \eta\,\mathrm{KL}
\end{aligned}
\end{equation*}
where $\hat{e}$ and $\hat{s}$ denote the evidence and search queries in $\hat{\tau}$, $R_{\text{acc}}$ scores the final answer, $R_{\text{align}}$ compares the sampled trajectory with the synthesized trajectory, $R_{\text{gain}}$ checks whether the retrieved evidence supports the answer, $R_{\text{query}}$ evaluates query quality, $R_{\text{fmt}}$ checks the tool-use format, and $\rho$ penalizes verbosity and repeated retrieval loops. The weights $w_1,\ldots,w_4$ balance these terms. $\mathrm{KL}$ is the GRPO regularization term and $\eta$ controls its weight. 

\begin{figure}[t]
    \centering
    \resizebox{\columnwidth}{!}{
    \begin{tikzpicture}[
        font=\sffamily,
        cat/.style={font=\bfseries\scriptsize, anchor=west},
        cnt/.style={font=\bfseries\tiny, anchor=west, text=black!78},
        seg/.style={draw=white, line width=0.35pt},
        barlabel/.style={font=\bfseries\tiny, text=white, align=center},
        darklabel/.style={font=\bfseries\tiny, text=black!75, align=center},
        smallbarlabel/.style={font=\bfseries\tiny, text=white, align=center, scale=0.82},
        smalldarklabel/.style={font=\bfseries\tiny, text=black!75, align=center, scale=0.82}
    ]
    \definecolor{cRet}{HTML}{3B6EA8}
    \definecolor{cUnd}{HTML}{4C956C}
    \definecolor{cRea}{HTML}{E58A2A}
    \definecolor{cGen}{HTML}{D45B5B}
    \definecolor{cCon}{HTML}{8E6BAE}

    \draw[-{Stealth[length=2.0mm]}, line width=0.45pt, black!55] (-0.18,0.10) -- (-0.18,3.95);
    \node[rotate=90, font=\bfseries\tiny, text=black!65] at (-0.31,2.00) {Increasing difficulty};
    \node[font=\bfseries\tiny, text=black!70, anchor=west] at (1.72,4.08) {Subtask distribution within each category};

    \node[cat, text=cRet, anchor=east] at (1.62,0.32) {Retrieval};
    \node[cnt, anchor=east] at (1.62,0.11) {1,300 \, (6.9\%)};
    \fill[seg, cRet!78] (1.72,0.04) rectangle ++(2.49,0.38);
    \fill[seg, cRet!48] (4.21,0.04) rectangle ++(3.99,0.38);
    \node[barlabel] at (2.97,0.23) {Clause Recall};
    \node[darklabel] at (6.21,0.23) {Clause Retrieval};

    \node[cat, text=cUnd, anchor=east] at (1.62,1.17) {Understanding};
    \node[cnt, anchor=east] at (1.62,0.96) {7,423 \, (39.2\%)};
    \fill[seg, cUnd!85] (1.72,0.89) rectangle ++(0.09,0.38);
    \fill[seg, cUnd!78] (1.81,0.89) rectangle ++(0.26,0.38);
    \fill[seg, cUnd!70] (2.07,0.89) rectangle ++(0.87,0.38);
    \fill[seg, cUnd!63] (2.94,0.89) rectangle ++(0.41,0.38);
    \fill[seg, cUnd!56] (3.35,0.89) rectangle ++(0.87,0.38);
    \fill[seg, cUnd!49] (4.22,0.89) rectangle ++(0.87,0.38);
    \fill[seg, cUnd!42] (5.09,0.89) rectangle ++(0.87,0.38);
    \fill[seg, cUnd!35] (5.96,0.89) rectangle ++(0.87,0.38);
    \fill[seg, cUnd!28] (6.83,0.89) rectangle ++(0.06,0.38);
    \fill[seg, cUnd!45] (6.89,0.89) rectangle ++(0.87,0.38);
    \fill[seg, cUnd!30] (7.76,0.89) rectangle ++(0.44,0.38);
    \node[smallbarlabel] at (2.51,1.08) {Concept};
    \node[smalldarklabel] at (3.79,1.08) {Subject};
    \node[smalldarklabel] at (4.66,1.08) {Reading};
    \node[smalldarklabel] at (5.53,1.08) {Event};
    \node[smalldarklabel] at (6.40,1.08) {Trigger};
    \node[smalldarklabel] at (7.33,1.08) {Opinion};

    \node[cat, text=cRea, anchor=east] at (1.62,2.02) {Reasoning};
    \node[cnt, anchor=east] at (1.62,1.81) {4,707 \, (24.9\%)};
    \fill[seg, cRea!82] (1.72,1.74) rectangle ++(0.64,0.38);
    \fill[seg, cRea!74] (2.36,1.74) rectangle ++(0.59,0.38);
    \fill[seg, cRea!66] (2.95,1.74) rectangle ++(0.56,0.38);
    \fill[seg, cRea!58] (3.51,1.74) rectangle ++(0.59,0.38);
    \fill[seg, cRea!50] (4.10,1.74) rectangle ++(1.38,0.38);
    \fill[seg, cRea!42] (5.48,1.74) rectangle ++(1.38,0.38);
    \fill[seg, cRea!34] (6.86,1.74) rectangle ++(1.38,0.38);
    \node[barlabel] at (4.79,1.93) {Scenario};
    \node[darklabel] at (6.17,1.93) {Legal Calc.};
    \node[darklabel] at (7.55,1.93) {Multi-hop};

    \node[cat, text=cGen, anchor=east] at (1.62,2.87) {Generation};
    \node[cnt, anchor=east] at (1.62,2.66) {2,928 \, (15.5\%)};
    \fill[seg, cGen!82] (1.72,2.59) rectangle ++(2.22,0.38);
    \fill[seg, cGen!68] (3.94,2.59) rectangle ++(0.93,0.38);
    \fill[seg, cGen!58] (4.87,2.59) rectangle ++(0.02,0.38);
    \fill[seg, cGen!47] (4.89,2.59) rectangle ++(2.22,0.38);
    \fill[seg, cGen!35] (7.11,2.59) rectangle ++(1.11,0.38);
    \node[barlabel] at (2.83,2.78) {Doc Sum.};
    \node[darklabel] at (6.00,2.78) {Opinion Sum.};
    \node[darklabel] at (7.67,2.78) {Trans.};

    \node[cat, text=cCon, anchor=east] at (1.62,3.72) {Consultation};
    \node[cnt, anchor=east] at (1.62,3.51) {2,582 \, (13.6\%)};
    \fill[seg, cCon!82] (1.72,3.44) rectangle ++(5.05,0.38);
    \fill[seg, cCon!58] (6.77,3.44) rectangle ++(0.40,0.38);
    \fill[seg, cCon!38] (7.17,3.44) rectangle ++(1.07,0.38);
    \node[barlabel] at (4.25,3.63) {User Consultation};
    \node[darklabel] at (7.71,3.63) {Judgment};
    \end{tikzpicture}}
    \caption{Distribution of the training data across major legal task categories and fine-grained subtasks. Detailed statistics are provided in Appendix~\ref{app:data_construction}.}
    \label{fig:training_data_distribution}
\end{figure}
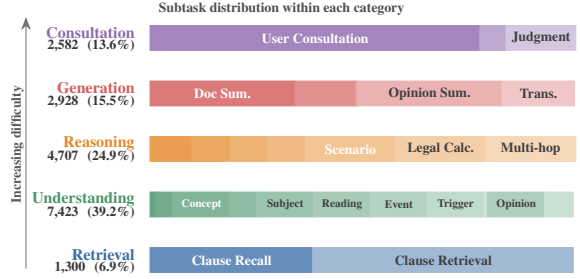

\section{Data Construction}
The dataset utilized in this project consists of two components: a legal knowledge base $\mathcal{K}$ and training data for legal agent $D$. 

\subsection{Legal Knowledge Base}\label{knowledge_base}

\begin{table*}[t]
  \centering
\small
\setlength{\tabcolsep}{3pt}
\scalebox{0.8}{%
      \begin{tabular}{lcc*{5}{c}} 
        \toprule
        \multirow{2}{*}{\textbf{Model}} & \multirow{2}{*}{\textbf{Retrieval}} & \multirow{2}{*}{\textbf{Scale}} & \multicolumn{2}{c}{\textbf{Lawbench}} & \multicolumn{2}{c}{\textbf{LexEval}} & \multirow{2}{*}{\textbf{Avg}} \\
        \cmidrule(lr){4-5} \cmidrule(lr){6-7}
        & & & \textbf{Trad. Score} & \textbf{LLM Judge} & \textbf{Trad. Score} & \textbf{LLM Judge} & \\ 
        \midrule
        LawGPT & \xmark & 7B  & 11.33 & 13.76 &9.49 & 9.18 & 10.94 \\
        Luwen& \xmark & 7B & 30.06 & 32.77 & 14.03 & 16.97&  23.45\\
         LegalOne& \xmark & 8B  & 23.25 & 41.02 & 35.30 & 43.79& 35.84 \\
        Qwen3-8B & \xmark & 8B & 24.70 & 40.20 & 18.04 & 28.47 & 27.85 \\
        A2search& \cmark & 7B  & 46.82 & 49.14 & 35.23 & 37.94 & 42.28 \\
        R-Search & \cmark & 7B & 43.01 & 47.44 & 45.05 & 46.85 & 45.59 \\

        Qwen3-8B-Embedding& \cmark & 8B  & 32.92 & 43.25 & 35.79 & 41.40 & 38.34  \\
        Llama-Embed-Nemotron& \cmark & 8B &32.23&42.28 & 36.45 & 41.77 & 38.19 \\
        DeepSeek-V4-Flash-Preview & \xmark & 284B  & 62.40 & 68.77 & 61.81 & 72.90& 66.47 \\
        GPT-5.4-Mini & \xmark & -- & 54.83 & 62.83 & 50.50 & 59.32& 56.87 \\
        Legal-R1  & \cmark & 8B   &57.36 & 63.97& 56.54 & 63.28 & 60.28 \\

        \bottomrule
      \end{tabular}
  }
    \caption{Model Performance Comparison on LexEval,and Lawbench Benchmarks. Judge by DeepSeek-V4-Flash-Preview.}
  \label{tab:main-result}
\end{table*}

\noindent\textbf{Normative Legal Corpus}.
We construct $\mathcal{K}_{norm}$ from the National Database of Laws and Regulations\footnote{\url{https://flk.npc.gov.cn/index}}. Each statutory article or judicial interpretation is treated as an individual retrieval unit and is annotated with its law title, chapter, section, article number, and effective date. The processed corpus contains 296 active laws with 43,117 articles and 555 judicial interpretations with 6,370 articles.

\noindent\textbf{Criminal Precedent Corpus}.
We construct $\mathcal{K}_{case}$ from LeCaRDv2\footnote{\url{https://github.com/THUIR/LeCaRDv2}}, which contains 53,277 criminal judgment records. Each case is retained as a precedent-level retrieval unit and is associated with structured attributes including the court level, case facts, judgment outcome, and sentencing result. This corpus supports precedent retrieval for analogous facts and sentencing comparison.

\noindent\textbf{Authority and Sentencing Annotations}.
To support authority-aware and diversity-aware precedent retrieval, we add two annotations to $\mathcal{K}_{case}$. First, cases are grouped by the level of the adjudicating court, including high people's courts, intermediate people's courts, and primary people's courts or other local tribunals, and each group is mapped to an authority score. Second, the Punishment Severity Index (PSI) maps heterogeneous sentencing outcomes into a numerical severity value. Main penalties are converted into day-equivalent values and adjusted by penalty type and suspension factors, while supplementary penalties are added as extra severity terms:
\begin{equation*}
\text{PSI}=D \cdot \gamma \cdot \phi + \sum S_{\text{sup}}
\end{equation*}
where $D$ is the day-equivalent main penalty, $\gamma$ is the penalty-type coefficient, $\phi$ is the suspension factor, and $S_{\text{sup}}$ denotes supplementary penalties such as fines or deprivation of political rights. Detailed authority scores and PSI examples are provided in Appendix~\ref{sec:appendix_kb}.








\subsection{Training Data Construction}
We construct the post-training dataset through an automated data synthesis and purification pipeline. The data are collected from established legal benchmarks and authoritative examination resources, including CAIL \citep{xiao2018cail2018}, DISC-LawLLM \citep{yue2023disc}, CrimeKgAssitant\footnote{\url{https://github.com/liuhuanyong/CrimeKgAssitant}} , and the National Unified Legal Professional Qualification Examination\footnote{\url{https://www.moj.gov.cn/jgsz/jgszzsdw/zsdwgjsfkszx/}}, and are further complemented with domain-specific synthetic problem sets. We then converts heterogeneous legal materials into task-specific QA pairs while applying multiple quality-control mechanisms. Specifically, the pipeline first performs task-driven applicability gating to filter source texts that are unsuitable for a given task, then removes potentially contaminated samples through $N$-gram fingerprint matching against held-out benchmarks, and finally applies automatic leakage detection and rule-based distractor construction to improve the reliability of open-ended and multiple-choice questions. The resulting dataset contains 18,940 high-quality structured QA instances, covering five major legal domains and 29 fine-grained subcategories. Figure~\ref{fig:training_data_distribution} shows the distribution of the training data across these categories. The detailed distribution and construction details are provided in Appendix~\ref{app:data_construction}.

\section{Experiments}

\subsection{Setup}
\noindent\textbf{Benchmark}. We evaluate our method on LexEval \citep{li2024lexeval} and LawBench \citep{fei2024lawbench}, two widely used Chinese legal benchmarks. LexEval consists of 23 tasks and 14,150 questions covering six legal cognitive abilities, while LawBench includes 20 tasks across three judicial cognitive dimensions, ranging from multiple-choice questions to practical applications such as reading comprehension, criminal amount calculation, and legal consultation.

\noindent\textbf{Evaluation Metrics}. 
We evaluate the proposed method from two perspectives: \textit{performance} and \textit{efficiency}. Performance is measured by benchmark default scores and an \textit{LLM-as-a-judge} score, while efficiency is assessed by retrieval turns, execution time, tool-invocation time, and token consumption.

\noindent\textbf{Implementation Details}. 
We train Legal-R1 from Qwen3-8B using SFT followed by GRPO. The SFT stage runs for 3 epochs on $D_{\text{SFT}}$ with a learning rate of $5\times10^{-6}$ and a batch size of 32. The GRPO stage uses a learning rate of $1\times10^{-6}$, a batch size of 32, 8 rollouts, and $\eta=0.08$. Training is conducted on four NVIDIA H20 GPUs. At inference time, Legal-R1 serves as the lawyer model that performs multi-turn reasoning and retrieval. A hybrid retriever and an auxiliary Qwen3-8B assistant model refine the retrieved evidence, while another Qwen3-8B judge model generates the final evidence-grounded answer. Full deployment details are provided in Appendix~\ref{app:implementation}.

\begin{figure}[t!] 
    \centering 
    \includegraphics[width=0.48\textwidth]{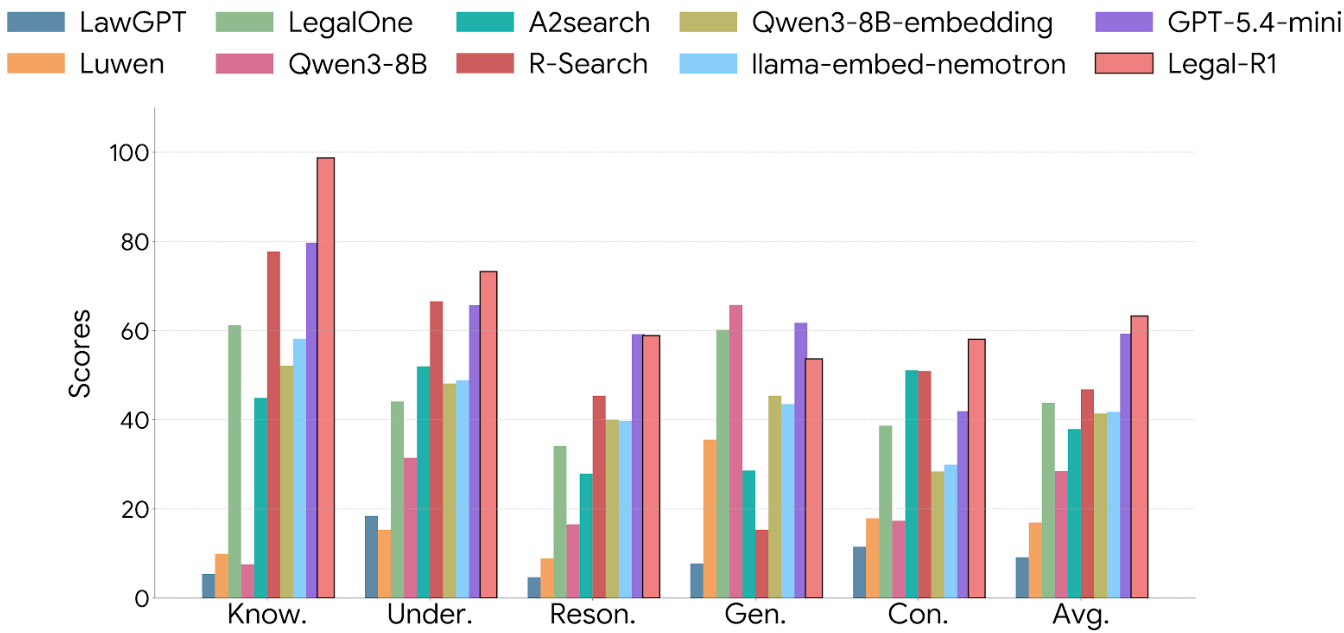} 
\caption{Model Performance across Five Categories of LexEval.}
    \label{fig:lexeval_category_performance} 
\end{figure}

\noindent\textbf{Baselines}.
We compare our model against three categories of baseline LLMs: (1) General LLMs, represented by DeepSeek-V4-Flash-Preview \citep{deepseekv4pro2026},and GPT-5.4-Mini \citep{openai2026gpt54mini}; (2) Legal-Specific LLMs, including LawGPT \citep{zhou2024lawgpt},  LegalOne \citep{li2026legalone}, and Luwen \citep{wu2026luwen}; and (3) RAG-based LLMs, which include A2search\citep{zhang20252} and R-Search \citep{zhao2025r}.


During evaluation, we configure each baseline LLM with its native prompt template (if available). For RAG-enabled baselines, we integrate their respective RAG components and load the same retrieval knowledge base used by our agentic retrieval framework. We then execute a unified evaluation pipeline: the primary models and the judge model (Qwen3-8B) are deployed via vLLM to generate responses across all benchmarks and compute the final scores. Concurrently, we track key inference metrics, including the number of retrieval turns, execution time (latency), and token consumption.

\subsection{Results and Discussion}
\begin{figure}[t!] 
    \centering 
    \includegraphics[width=0.48\textwidth]{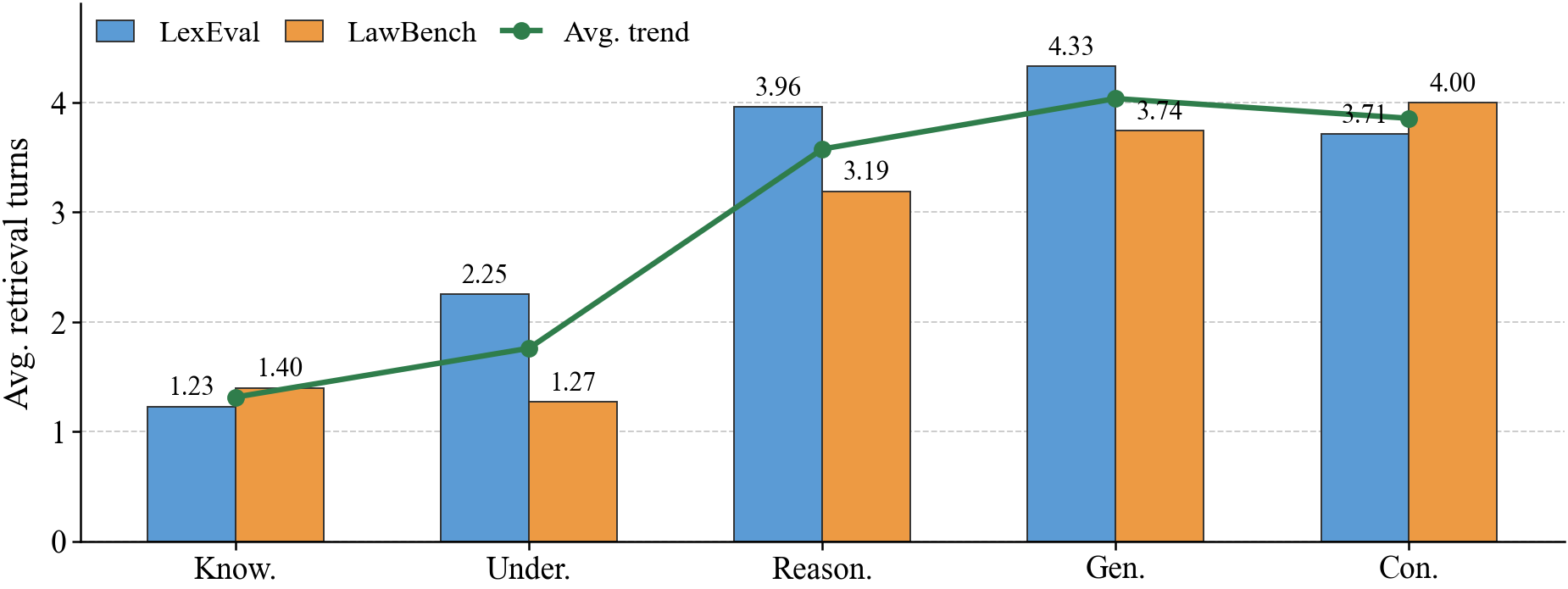} 
    \caption{Average Retrieval Turn for Different Complexity Tasks on Lawbench and Lexeval Benchmark.} 
    \label{fig:avg_turn} 
\end{figure}

\begin{table*}[t]
  \centering
    \small
    \setlength{\tabcolsep}{3pt}
    \scalebox{0.8}{%
      \begin{tabular}{lcc*{5}{c}} 
        \toprule
        \multirow{2}{*}{\textbf{Model}} & \multirow{2}{*}{\textbf{Retrieval}} & \multirow{2}{*}{\textbf{Scale}} & \multicolumn{2}{c}{\textbf{Lawbench}} & \multicolumn{2}{c}{\textbf{LexEval}} & \multirow{2}{*}{\textbf{Avg}} \\
        \cmidrule(lr){4-5} \cmidrule(lr){6-7}
        & & & \textbf{Trad. Score} & \textbf{LLM Judge} & \textbf{Trad. Score} & \textbf{LLM Judge} & \\ 
        \midrule
        Qwen3-8B & \xmark & 8B & 24.70 & 40.20 & 18.04 & 28.47 & 27.85 \\
        Qwen3-8B w/ post-train  & \xmark & 8B   &26.24 & 38.98 & 55.51 & 62.78 & 45.88 \\
        Qwen3-8B w/ hybrid      & \cmark & 8B   & 52.02 & 59.85 &50.52 & 61.49 &55.91\\
        Qwen3-8B w/ rerank      & \cmark & 8B   &53.28 & 57.97 & 50.12 & 57.31 & 54.67 \\
        Legal-R1  w/ hybrid & \cmark & 8B  &57.08& 63.69 & 55.47 & 61.96 &59.55\\
        Legal-R1 w/ rerank  & \cmark & 8B   &57.36 & 63.97& 56.54 & 63.28 & 60.28 \\
        \bottomrule
      \end{tabular}
  }
    \caption{Ablation study: Model Performance Comparison on LexEval, Lawbench and average Benchmarks. Judge by DeepSeek-V4-Flash-Preview.}
  \label{tab:ablation_study}
\end{table*}

\begin{table*}[t]
\centering
\small
\setlength{\tabcolsep}{3pt}
\scalebox{0.8}{%
\begin{tabular}{p{0.1\linewidth}p{0.9\linewidth}}
\toprule
\multicolumn{2}{l}{\textbf{Case Study: Autonomous Multi-Turn Retrieval}} \\
\midrule
\textbf{Query} & Will a debtor automatically be blacklisted if they fail to repay a debt within the court's 25-day deadline? \\
\textbf{Trajectory} &
\faLightbulb~\textbf{Thinking Step 1:} I need to locate the foundational statutory criteria for blacklisting. $\hookrightarrow$ \faSearch~\textbf{Retrieval 1:} \textit{Article 1}.$\hookrightarrow$ 
\faLightbulb~\textbf{Thinking Step 2:} Article 1 hinges on "fulfillment capability." I must retrieve its specific definition. $\hookrightarrow$ \faSearch~\textbf{Retrieval 2:} \textit{Article 2}. $\hookrightarrow$
\faLightbulb~\textbf{Thinking Step 3:} I need to search for procedural exemptions regarding the blacklisting process. $\hookrightarrow$ \faSearch~\textbf{Retrieval 3:} \textit{Article 3}. $\hookrightarrow$
\faLightbulb~\textbf{Thinking Step 4:} Let me invoke a precedent search to validate how courts apply these articles in practice.$\hookrightarrow$ \faSearch~\textbf{Retrieval 4:} \textit{Precedent Case}.
\\
\textbf{Answer} &
No, blacklisting is not automatic; according to Articles 1, 2, and 3, it strictly depends on whether the debtor willfully refuses to comply despite having the financial capability, provided they do not meet any legal exemptions. {\faThumbsUp} \\
\bottomrule
\end{tabular}
}
\caption{A case study of Legal-R1's autonomous multi-turn retrieval, with the full trajectory provided in Table~\ref{fig:case_study_full} in the appendix.}
\label{tab:case_study}
\end{table*}

\paragraph{Main Results}
We evaluate various baselines on legal tasks across Lawbench and LexEval, as shown in Table~\ref{tab:main-result}. The empirical results indicate that while traditional legal-specific models and RAG-augmented methods show competitive performance, they are consistently outperformed by our approach. Our method, Legal-R1, performs exceptionally well across all datasets and evaluation metrics, achieving a top-tier overall average score of 60.28\%. 
At the 8B parameter scale, Legal-R1 outperforms the strongest RAG baseline, R-Search, by 14.69 points in overall average and exceeds GPT-5.4-Mini by 3.41 points. Legal-R1 ranks second to DeepSeek-V4-Flash-Preview (284B). Figure~\ref{fig:lexeval_category_performance} further presents the category-level results on the five LexEval categories, providing a more fine-grained view of Legal-R1's performance across different dimensions of legal reasoning. Overall, these results demonstrate the strong and broadly effective legal-reasoning capability of Legal-R1.

\paragraph{Efficiency}
As illustrated in Figure~\ref{fig:avg_turn}, retrieval frequency varies with task complexity: the system uses an average of 1.2--1.7 turns for \textit{Knowledge} and \textit{Understanding} tasks, and 3.1--4.3 turns for \textit{Reasoning}, \textit{Generation}, and \textit{Consultation} tasks. This pattern is observed on both LawBench and LexEval. The additional retrieval has a measurable cost. Relative to Qwen3-8B w/ hybrid, Legal-R1 raises the overall average score from 55.91 to 60.28, while the average LawBench retrieval/reasoning turns increase from 0.35 to 2.33 and LawBench token consumption increases from 5,196.62 to 10,861.85. The score is averaged across benchmarks, whereas the cost measures are from LawBench; the complete comparison is reported in Appendix~\ref{app:efficiency}.

\paragraph{Ablation Study}

To verify the effectiveness of each component in our framework, we conduct an ablation study on Lawbench and LexEval, as reported in Table~\ref{tab:ablation_study}. 

First, the results underscore that incorporating external knowledge is paramount for legal reasoning; equipping the vanilla Qwen3-8B with hybrid retrieval (`w/ hybrid') drastically boosts the overall average score from 27.85\% to 55.91\%. Second, domain-specific adaptation without retrieval (`w/ post-train') also yields a substantial improvement (+18.03 points on average), validating the necessity of legal domain fine-tuning. Finally, our complete method, Legal-R1 , integrates domain post-training, the multi-stage RAG pipeline, and GRPO reinforcement learning to achieve the highest average score of 60.28\%. This final leap demonstrates that these components provide orthogonal benefits, culminating in optimal synergistic performance in complex legal scenarios.

\noindent\textbf{Adaptive retrieval and evidence processing.}
Table~\ref{tab:lawbench_additional_ablation} reports LawBench-only ablations. The fixed-round variants modify the prompt and retrieval procedure to require one or three retrieval rounds, whereas the complete system plans retrieval actions and decides when to stop.

\begin{table}[t]
\centering
\small
\begin{tabular}{lrr}
\toprule
\textbf{Legal-R1 Pipeline} & \textbf{LLM Judge} & \textbf{Trad. Score} \\
\midrule
Fixed 1-turn retrieval & 61.70 & 53.95 \\
Fixed 3-turn retrieval & 62.75 & 54.65 \\
W/ hybrid & 63.69 & 57.08 \\
W/o LLM rerank & 62.99 & 57.35 \\
W/o summarization & 61.44 & 52.95 \\
Full & 63.97 & 57.36 \\
\bottomrule
\end{tabular}
\caption{LawBench ablations of retrieval rounds and evidence-processing components.}
\label{tab:lawbench_additional_ablation}
\end{table}

The complete system scores 57.36 on the traditional metric, compared with 53.95 and 54.65 for fixed one-turn and three-turn retrieval, respectively. These comparisons support adaptive retrieval planning, although the fixed-round variants also change the prompt and retrieval procedure. Removing evidence summarization reduces the traditional score to 52.95 and the LLM-judge score to 61.44. Removing LLM reranking has little effect on the traditional score (57.35 versus 57.36) but reduces the LLM-judge score from 63.97 to 62.99.

\noindent\textbf{Statute retrieval quality.}
We evaluate retrieval on 769 provision-oriented queries; DeepSeek-V3 synthesizes verifiable QA pairs from legal benchmark materials and provisions. The corresponding answers are associated with provisions in the retrieval knowledge base. Table~\ref{tab:retrieval_quality} compares the resulting retrieval rankings with text2vec and BM25.

\begin{table}[t]
\centering
\small
\begin{tabular}{lrrrr}
\toprule
\textbf{System} & \textbf{nDCG} & \textbf{Recall} & \textbf{MRR} & \textbf{Hit Rate} \\
\midrule
text2vec & 0.3187 & 0.4109 & 0.2884 & 0.4109 \\
BM25 & 0.6993 & 0.7789 & 0.6725 & 0.7789 \\
Legal-R1 & 0.7659 & 0.8088 & 0.7516 & 0.8088 \\
\bottomrule
\end{tabular}
\caption{Statute retrieval quality on 769 provision-oriented queries.}
\label{tab:retrieval_quality}
\end{table}

Relative to BM25, Legal-R1 improves nDCG from 0.6993 to 0.7659 and MRR from 0.6725 to 0.7516; recall and hit rate also rise from 0.7789 to 0.8088. These results provide direct evidence of improved retrieval ranking for provision-oriented requests. 

\paragraph{Case Study}
To intuitively illustrate the autonomous multi-turn reasoning and retrieval capabilities of Legal-R1, we conduct a qualitative case study on a representative real-world legal consultation. As analyzing in Table~\ref{tab:case_study}, given a user query regarding enforcement penalties, the model does not rely on static memories but independently initiates a sequence of precise retrieval loops to search for statutory criteria and validation procedures for dishonest judgment debtors. Notably, it strategically executes cross-references between general rules (Article 1) and exception clauses (Article 3), while invoking a legal precedent to ground its contextual understanding in judicial practice. Ultimately, through this systematic deliberation, Legal-R1 synthesizes an accurate and legally sound judgment, demonstrating its robust reliability in professional legal applications.

\begin{figure*}[t]
\begin{tcolorbox}[title={Case Study: Autonomous Multi-Turn Retrieval and Reasoning Trajectory}]
\textbf{[User Query]} \\
My friend borrowed 164,000 RMB this January, failed to repay, and was sued. We won the lawsuit, and the court ordered repayment within 25 days. If they still do not pay, will they automatically be blacklisted as a dishonest judgment debtor?

\vspace{0.5em}
\textbf{[Autonomous Reasoning \& Retrieval Trajectory]} \\
\textit{Thinking Step 1:} To address whether non-repayment automatically leads to being blacklisted, I must first locate the precise statutory criteria and procedures regarding the designation of a "dishonest judgment debtor." \\
$\hookrightarrow$ \textbf{Retrieval 1 (Statute):} \textit{Several Provisions of the Supreme People's Court on Publishing Information of the List of Dishonest Judgment Debtors, Article 1.} \\
\textit{Thinking Step 2:} According to Article 1, "having the capability to fulfill but refusing to fulfill" is a core mandatory prerequisite. I need to retrieve further regulatory interpretations on how the court determines "the capability to fulfill." \\
$\hookrightarrow$ \textbf{Retrieval 2 (Statute):} \textit{Several Provisions of the Supreme People's Court on Publishing Information of the List of Dishonest Judgment Debtors, Article 2.} \\
\textit{Thinking Step 3:} I must now locate procedural guidelines or exemption clauses, particularly regarding situations where execution remains incomplete after the court-ordered deadline. \\
$\hookrightarrow$ \textbf{Retrieval 3 (Statute):} \textit{Several Provisions of the Supreme People's Court on Publishing Information of the List of Dishonest Judgment Debtors, Article 3.} \\
\textit{Thinking Step 4:} To validate the interactive boundaries of "non-fulfillment" and "blacklisting" in real judicial practice, I will invoke a precedent retrieval to observe how local courts adjudicate analogous scenarios. \\
$\hookrightarrow$ \textbf{Retrieval 4 (Precedent Case):} \textit{The judgment debtor Xiang Mo failed to fulfill the repayment obligation after the court judgment took effect, and still willfully refused execution after receiving the enforcement notice...}

\vspace{0.5em}
\textbf{[Final Response Generated by Judge Agent]} \\
\textbf{Answer}: No, failure to repay within 25 days does not inevitably lead to being blacklisted. The critical criterion hinges on whether the debtor "has the capability to fulfill but willfully refuses to do so." If the debtor genuinely lacks the capacity to repay (e.g., assets are already sealed by other creditors, or income is insufficient for basic living costs), they may not be blacklisted. Furthermore, according to Article 3 of the \textit{Provisions}, if the debtor provides sufficient collateral, or if their seized/frozen assets are already adequate to liquidate the debt, they shall not be blacklisted. Therefore, the court evaluates this on a case-by-case basis. \\
\textbf{Legal Basis}: Articles 1 and 3 of the \textit{Several Provisions of the Supreme People's Court on Publishing Information of the List of Dishonest Judgment Debtors}.
\end{tcolorbox}
\caption{An idealized snapshot of Legal-R1's autonomous thought-retrieval loop and final response synthesis.}
\label{fig:case_study_full}
\end{figure*}

\paragraph{SFT vs RL}

\begin{table}[t]
  \centering
    \small
    \setlength{\tabcolsep}{3pt}
    \scalebox{0.8}{%
      \begin{tabular}{lcc}
        \toprule
        \textbf{Model} & \textbf{Retrieval} & \textbf{Avg. Score} \\ 
        \midrule
        Qwen3-8B      & \cmark & 54.67 \\
        Qwen3-8B-SFT  & \cmark & 59.98 \\
        Legal-R1               & \cmark & 60.28\\
        \bottomrule
      \end{tabular}
  }
  \caption{Ablation study of training progression based on overall average scores.}
  \label{tab:training_ablation}
\end{table}

We isolate the distinct contributions of supervised fine-tuning (SFT) and reinforcement learning (RL) within our framework, as presented in Table~\ref{tab:training_ablation}. Transitioning from the raw RAG pipeline (`w/ rerank') to the SFT stage (`Qwen3-8B-SFT w/ rerank') yields a prominent performance leap of 5.31 percentage points, demonstrating that SFT is vital for injecting foundational legal expertise and task-specific alignment. Building upon this optimized baseline, Legal-R1 leverages GRPO reinforcement learning to further elevate the overall average score to a peak of 60.28\%. This continuous gain validates that RL successfully refines the model's underlying reasoning policy, capitalizing on the specialized knowledge acquired during SFT to unlock maximum precision.

\section{Conclusion}
In this paper, we introduced Legal-R1, an agentic legal large language model for evidence-based question answering. Our framework couples a multi-turn Lawyer Agent with a multi-stage RAG pipeline via a Legal Assistant Agent. This pipeline uses hybrid retrieval followed by authority- and diversity-aware reranking, and leverages LLM filtering and summarization to distill compact legal evidence units. To optimize the central retrieval policy, Legal-R1 undergoes a streamlined two-stage training paradigm of Evidence-Masked Supervised Fine-Tuning (SFT) and Trajectory-Aware Reinforcement Learning via GRPO. Empirical results on the LexEval and LawBench benchmarks demonstrate the exceptional efficacy and efficiency of our approach. At the 8B parameter scale, Legal-R1 achieves a state-of-the-art overall average score of 60.28\%, surpassing the proprietary GPT-5.4-Mini. Furthermore, our model maintains high efficiency by dynamically scaling its retrieval turns according to task complexity, achieving a favorable balance between computational cost and performance.

\section*{Limitations}
While Legal-R1 demonstrates robust and highly accurate performance in autonomous legal reasoning, it is not without minor limitations. Primarily, the multi-turn agentic retrieval process---although dynamically adjusted for task complexity---inherently introduces a slight inference latency compared to standard, single-pass generation models. While this computational overhead is justified by the significant gains in accuracy and legal grounding, it may require further pipeline optimization (e.g., parallelized evidence fetching or KV-cache optimization) before deployment in extremely time-sensitive, real-time legal consultation environments. Additionally, the current model and knowledge base are deeply tailored to the Chinese legal system; extending this framework to support multi-jurisdictional or multilingual legal reasoning remains a promising avenue for future work.

\bibliography{custom}

@article{shao2024deepseekmath,
  title={Deepseekmath: Pushing the limits of mathematical reasoning in open language models},
  author={Shao, Zhihong and Wang, Peiyi and Zhu, Qihao and Xu, Runxin and Song, Junxiao and Bi, Xiao and Zhang, Haowei and Zhang, Mingchuan and Li, YK and Wu, Yang and others},
  journal={arXiv preprint arXiv:2402.03300},
  year={2024}
}

@inproceedings{carbonell1998use,
  title={The use of MMR, diversity-based reranking for reordering documents and producing summaries},
  author={Carbonell, Jaime and Goldstein, Jade},
  booktitle={Proceedings of the 21st annual international ACM SIGIR conference on Research and development in information retrieval},
  pages={335--336},
  year={1998}
}

@article{li2024lexeval,
  title={Lexeval: A comprehensive chinese legal benchmark for evaluating large language models},
  author={Li, Haitao and Chen, You and Ai, Qingyao and Wu, Yueyue and Zhang, Ruizhe and Liu, Yiqun},
  journal={Advances in Neural Information Processing Systems},
  volume={37},
  pages={25061--25094},
  year={2024}
}

@inproceedings{fei2024lawbench,
  title={Lawbench: Benchmarking legal knowledge of large language models},
  author={Fei, Zhiwei and Shen, Xiaoyu and Zhu, Dawei and Zhou, Fengzhe and Han, Zhuo and Huang, Alan and Zhang, Songyang and Chen, Kai and Yin, Zhixin and Shen, Zongwen and others},
  booktitle={Proceedings of the 2024 conference on empirical methods in natural language processing},
  pages={7933--7962},
  year={2024}
}

@misc{deepseekv4pro2026,
  author       = {{DeepSeek-AI}},
  title        = {{DeepSeek-V4-Pro}},
  howpublished = {\url{https://huggingface.co/deepseek-ai/DeepSeek-V4-Pro}},
  year         = {2026},
  note         = {Accessed: 2026-05-25}
}

@misc{openai2026gpt54mini,
  author       = {{OpenAI}},
  title        = {{Introducing GPT-5.4 mini and nano}},
  howpublished = {\url{https://openai.com/index/introducing-gpt-5-4-mini-and-nano/}},
  year         = {2026},
  note         = {Accessed: 2026-05-25}
}

@article{zhou2024lawgpt,
  title={Lawgpt: A chinese legal knowledge-enhanced large language model},
  author={Zhou, Zhi and Shi, Jiang-Xin and Song, Peng-Xiao and Yang, Xiao-Wen and Jin, Yi-Xuan and Guo, Lan-Zhe and Li, Yu-Feng},
  journal={arXiv preprint arXiv:2406.04614},
  year={2024}
}

@article{li2026legalone,
  title={LegalOne: a family of foundation models for reliable legal reasoning},
  author={Li, Haitao and Chen, Yifan and Miao, Shuo and Dong, Qian and Chen, Jia and Hu, Yiran and Chen, Junjie and Qin, Minghao and Wu, Yueyue and Zhou, Yujia and others},
  journal={arXiv preprint arXiv:2602.00642},
  year={2026}
}

@article{wu2026luwen,
  title={Luwen Technical Report},
  author={Wu, Yiquan and Liu, Yuhang and Liu, Yifei and Li, Ang and Zhou, Siying and Kuang, Kun},
  journal={arXiv preprint arXiv:2604.06737},
  year={2026}
}

@article{zhang20252,
  title={{$A^{2}$} Search: Ambiguity-Aware Question Answering with Reinforcement Learning},
  author={Zhang, Fengji and Niu, Xinyao and Ying, Chengyang and Lin, Guancheng and Hao, Zhongkai and Fan, Zhou and Huang, Chengen and Keung, Jacky and Chen, Bei and Lin, Junyang},
  journal={arXiv preprint arXiv:2510.07958},
  year={2025}
}

@article{zhao2025r,
  title={R-search: Empowering llm reasoning with search via multi-reward reinforcement learning},
  author={Zhao, Qingfei and Wang, Ruobing and Xu, Dingling and Zha, Daren and Liu, Limin},
  journal={arXiv preprint arXiv:2506.04185},
  year={2025}
}

@article{xiao2018cail2018,
  title={Cail2018: A large-scale legal dataset for judgment prediction},
  author={Xiao, Chaojun and Zhong, Haoxi and Guo, Zhipeng and Tu, Cunchao and Liu, Zhiyuan and Sun, Maosong and Feng, Yansong and Han, Xianpei and Hu, Zhen and Wang, Heng and others},
  journal={arXiv preprint arXiv:1807.02478},
  year={2018}
}

@article{yue2023disc,
  title={Disc-lawllm: Fine-tuning large language models for intelligent legal services},
  author={Yue, Shengbin and Chen, Wei and Wang, Siyuan and Li, Bingxuan and Shen, Chenchen and Liu, Shujun and Zhou, Yuxuan and Xiao, Yao and Yun, Song and Huang, Xuanjing and others},
  journal={arXiv preprint arXiv:2309.11325},
  year={2023}
}

@article{lewis2020retrieval,
  title={Retrieval-augmented generation for knowledge-intensive nlp tasks},
  author={Lewis, Patrick and Perez, Ethan and Piktus, Aleksandra and Petroni, Fabio and Karpukhin, Vladimir and Goyal, Naman and K{\"u}ttler, Heinrich and Lewis, Mike and Yih, Wen-tau and Rockt{\"a}schel, Tim and others},
  journal={Advances in neural information processing systems},
  volume={33},
  pages={9459--9474},
  year={2020}
}

@inproceedings{karpukhin2020dense,
  title={Dense passage retrieval for open-domain question answering},
  author={Karpukhin, Vladimir and Oguz, Barlas and Min, Sewon and Lewis, Patrick and Wu, Ledell and Edunov, Sergey and Chen, Danqi and Yih, Wen-tau},
  booktitle={Proceedings of the 2020 conference on empirical methods in natural language processing (EMNLP)},
  pages={6769--6781},
  year={2020}
}

@inproceedings{sun2023chatgpt,
  title={Is ChatGPT good at search? investigating large language models as re-ranking agents},
  author={Sun, Weiwei and Yan, Lingyong and Ma, Xinyu and Wang, Shuaiqiang and Ren, Pengjie and Chen, Zhumin and Yin, Dawei and Ren, Zhaochun},
  booktitle={Proceedings of the 2023 conference on empirical methods in natural language processing},
  pages={14918--14937},
  year={2023}
}

@inproceedings{nogueira2020document,
  title={Document ranking with a pretrained sequence-to-sequence model},
  author={Nogueira, Rodrigo and Jiang, Zhiying and Pradeep, Ronak and Lin, Jimmy},
  booktitle={Findings of the association for computational linguistics: EMNLP 2020},
  pages={708--718},
  year={2020}
}

@article{gao2023retrieval,
  title={Retrieval-augmented generation for large language models: A survey},
  author={Gao, Yunfan and Xiong, Yun and Gao, Xinyu and Jia, Kangxiang and Pan, Jinliu and Bi, Yuxi and Dai, Yixin and Sun, Jiawei and Wang, Haofen and Wang, Haofen and others},
  journal={arXiv preprint arXiv:2312.10997},
  volume={2},
  number={1},
  pages={32},
  year={2023}
}

@article{yao2022react,
  title={React: Synergizing reasoning and acting in language models},
  author={Yao, Shunyu and Zhao, Jeffrey and Yu, Dian and Du, Nan and Shafran, Izhak and Narasimhan, Karthik and Cao, Yuan},
  journal={arXiv preprint arXiv:2210.03629},
  year={2022}
}

@article{schick2023toolformer,
  title={Toolformer: Language models can teach themselves to use tools},
  author={Schick, Timo and Dwivedi-Yu, Jane and Dess{\`\i}, Roberto and Raileanu, Roberta and Lomeli, Maria and Hambro, Eric and Zettlemoyer, Luke and Cancedda, Nicola and Scialom, Thomas},
  journal={Advances in neural information processing systems},
  volume={36},
  pages={68539--68551},
  year={2023}
}

@article{guo2025deepseek,
  title={Deepseek-r1: Incentivizing reasoning capability in llms via reinforcement learning},
  author={Guo, Daya and Yang, Dejian and Zhang, Haowei and Song, Junxiao and Wang, Peiyi and Zhu, Qihao and Xu, Runxin and Zhang, Ruoyu and Ma, Shirong and Bi, Xiao and others},
  journal={arXiv preprint arXiv:2501.12948},
  year={2025}
}

@article{cobbe2021training,
  title={Training verifiers to solve math word problems},
  author={Cobbe, Karl and Kosaraju, Vineet and Bavarian, Mohammad and Chen, Mark and Jun, Heewoo and Kaiser, Lukasz and Plappert, Matthias and Tworek, Jerry and Hilton, Jacob and Nakano, Reiichiro and others},
  journal={arXiv preprint arXiv:2110.14168},
  year={2021}
}

@article{hendrycks2021measuring,
  title={Measuring mathematical problem solving with the math dataset},
  author={Hendrycks, Dan and Burns, Collin and Kadavath, Saurav and Arora, Akul and Basart, Steven and Tang, Eric and Song, Dawn and Steinhardt, Jacob},
  journal={arXiv preprint arXiv:2103.03874},
  year={2021}
}

@article{ouyang2022training,
  title={Training language models to follow instructions with human feedback},
  author={Ouyang, Long and Wu, Jeffrey and Jiang, Xu and Almeida, Diogo and Wainwright, Carroll and Mishkin, Pamela and Zhang, Chong and Agarwal, Sandhini and Slama, Katarina and Ray, Alex and others},
  journal={Advances in neural information processing systems},
  volume={35},
  pages={27730--27744},
  year={2022}
}

@inproceedings{wang2023self,
  title={Self-instruct: Aligning language models with self-generated instructions},
  author={Wang, Yizhong and Kordi, Yeganeh and Mishra, Swaroop and Liu, Alisa and Smith, Noah A and Khashabi, Daniel and Hajishirzi, Hannaneh},
  booktitle={Proceedings of the 61st annual meeting of the association for computational linguistics (volume 1: long papers)},
  pages={13484--13508},
  year={2023}
}

@article{xu2024benchmark,
  title={Benchmark data contamination of large language models: A survey},
  author={Xu, Cheng and Guan, Shuhao and Greene, Derek and Kechadi, M and others},
  journal={arXiv preprint arXiv:2406.04244},
  year={2024}
}

@article{xiao2026structuring,
  title={Structuring Human Objectives: A Survey of Rubrics for Evaluation, Alignment, and Agentic AI},
  author={Xiao, Hongru and Li, Jie and Li, Zhirui and Yang, Bin and Li, Xiang and Li, Sunzhu and Han, Jiale},
  journal={SSRN},
  year={2026},
  doi={10.2139/ssrn.6970459},
  url={https://ssrn.com/abstract=6970459}
}

@inproceedings{wiratunga2024cbr,
  title={CBR-RAG: case-based reasoning for retrieval augmented generation in LLMs for legal question answering},
  author={Wiratunga, Nirmalie and Abeyratne, Ramitha and Jayawardena, Lasal and Martin, Kyle and Massie, Stewart and Nkisi-Orji, Ikechukwu and Weerasinghe, Ruvan and Liret, Anne and Fleisch, Bruno},
  booktitle={International Conference on Case-Based Reasoning},
  pages={445--460},
  year={2024},
  organization={Springer}
}

@inproceedings{louis2024interpretable,
  title={Interpretable long-form legal question answering with retrieval-augmented large language models},
  author={Louis, Antoine and Van Dijck, Gijs and Spanakis, Gerasimos},
  booktitle={Proceedings of the AAAI conference on artificial intelligence},
  volume={38},
  number={20},
  pages={22266--22275},
  year={2024}
}

@article{van2017concept,
  title={On the concept of relevance in legal information retrieval},
  author={Van Opijnen, Marc and Santos, Cristiana},
  journal={Artificial Intelligence and Law},
  volume={25},
  number={1},
  pages={65--87},
  year={2017},
  publisher={Springer}
}

@article{dahl2024large,
  title={Large legal fictions: Profiling legal hallucinations in large language models},
  author={Dahl, Matthew and Magesh, Varun and Suzgun, Mirac and Ho, Daniel E},
  journal={Journal of Legal Analysis},
  volume={16},
  number={1},
  pages={64--93},
  year={2024},
  publisher={Oxford University Press UK}
}

@inproceedings{asai2024self,
  title={Self-rag: Learning to retrieve, generate, and critique through self-reflection},
  author={Asai, Akari and Wu, Zeqiu and Wang, Yizhong and Sil, Avi and Hajishirzi, Hannaneh},
  booktitle={International conference on learning representations},
  volume={2024},
  pages={9112--9141},
  year={2024}
}

@inproceedings{li2024lecardv2,
  title={Lecardv2: A large-scale chinese legal case retrieval dataset},
  author={Li, Haitao and Shao, Yunqiu and Wu, Yueyue and Ai, Qingyao and Ma, Yixiao and Liu, Yiqun},
  booktitle={Proceedings of the 47th International ACM SIGIR Conference on Research and Development in Information Retrieval},
  pages={2251--2260},
  year={2024}
}

\appendix

\section{Prompts Used in the Agentic Retrieval Workflow}
\label{sec:appendix_prompts}

This appendix provides the prompts used by the LLM-based components in the proposed workflow. 
\subsection{Normative Evidence Reranking Prompt}
This prompt is used by the LLM-based normative reranker $\mathcal{L}_{norm}$. Given a retrieval intent and candidate legal provisions from $\mathcal{K}_{norm}$, the model ranks the candidates according to their relevance to the requested legal role, such as offense elements, sentencing ranges, or exception clauses. The detailed template implementation is illustrated in Figure~\ref{fig:normative_prompt}.

\begin{figure*}[t]
\begin{tcolorbox}[title={Normative Evidence Reranking Prompt}]
\#\#\# Task Instructions \\
You are an experienced legal document auditor. Based on the provided [Contextual Information] and [Search Query], please screen the [Candidate Texts] and select the document ID(s) that are semantically most relevant and strictly meet the statutory requirements.

\#\#\# Core Rules (Must Be Strictly Obeyed): \\
1. Context Priority: If the [Contextual Information] specifies a particular law name or article number (e.g., Article 2), the selected text must match it. If no matching text is found, return [] directly. \\
2. No Padding: The screening result should contain at most \textcolor{blue}{\{topk\}} segments. If only 1 segment matches, return only that segment's ID; if none match, return []. It is strictly forbidden to return irrelevant text just to reach the \textcolor{blue}{\{topk\}} limit. \\
3. Output Format: You are only allowed to output a Python-formatted list of integers, for example: [1] or [1, 2]. \\
4. No Explanation: Do not output any analytical process, thinking content ($<$think$>$), or any extra text explanations. \\
5. Index Alignment: Please carefully read the content of each candidate text. If the [Contextual Information] requests `Article 2', locate the text that actually describes `Article 2' and output the corresponding [ID] of that text.

\#\#\# Input Data \\
- [Search Query]: \textcolor{blue}{\{query\}} \\
- [Contextual Information]: \textcolor{blue}{\{context\}} \\
- [Candidate Texts]: \textcolor{blue}{\{results\}}

\#\#\# Screening Result (Output List Only): 
\end{tcolorbox}
\caption{The prompt template utilized for normative evidence reranking. Within this template, \textcolor{blue}{\{topk\}} denotes the predefined upper bound of retrieval turns, \textcolor{blue}{\{query\}} and \textcolor{blue}{\{context\}} represent the primary retrieval informational query, and \textcolor{blue}{\{results\}} contains the candidate document pool obtained from the initial coarse-grained retrieval stage.}
\label{fig:normative_prompt}
\end{figure*}

\subsection{Precedent Evidence Summarization Prompt}
This prompt is used by the case summarizer $\mathcal{L}_{case}$. Given a selected precedent, the model compresses it into an evidence unit that preserves the key facts, court level, judgment outcome, and sentencing information. The complete template configuration is detailed in Figure~\ref{fig:precedent_prompt}.

\begin{figure*}[t]
\begin{tcolorbox}[title={Precedent Evidence Summarization Prompt}]
\#\#\# Task Instructions \\
You are an experienced judicial assistant. Please carefully compare the user's [Queried Case Facts] with the retrieved [Candidate Case]. \\
Summarize the ``facts,'' ``judicial reasoning,'' and ``judgment outcome'' of the candidate case into a concise 150-word case brief.

\#\#\# Briefing Requirements: \\
1. Summarize the facts of the case, emphasizing the similarities between this case and the [Queried Case Facts]. \\
2. Concisely outline the court's judicial reasoning (especially the underlying logic regarding the determination of key circumstances). \\
3. Clearly state the final judgment outcome.

\#\#\# Input Data \\
- [Queried Case Facts]: \textcolor{blue}{\{fact\_query\}} \\
- [Candidate Case - Facts]: \textcolor{blue}{\{doc\_fact\}} \\
- [Candidate Case - Judicial Reasoning]: \textcolor{blue}{\{doc\_reason\}} \\
- [Candidate Case - Judgment Outcome]: \textcolor{blue}{\{doc\_result\}}

\#\#\# Output (Please directly output the 150-word brief):
\end{tcolorbox}
\caption{The prompt template utilized for precedent evidence summarization. Within this template, \textcolor{blue}{\{fact\_query\}} denotes the target case query to be investigated, while \textcolor{blue}{\{doc\_fact\}}, \textcolor{blue}{\{doc\_reason\}}, and \textcolor{blue}{\{doc\_result\}} represent the factual narrative, the judicial rationale, and the final adjudication outcome of the retrieved candidate precedent, respectively.}
\label{fig:precedent_prompt}
\end{figure*}

\subsection{Judge Answer Generation Prompt}
This prompt is used by the Judge agent $\mathcal{L}_{Judge}$. Given the legal question and the collected retrieval trajectory, the model generates the final answer based only on the retrieved legal norms and precedents, while avoiding unsupported claims. The structured implementation of this template is presented in Figure~\ref{fig:judge_prompt}.

\begin{figure*}[t]
\begin{tcolorbox}[title={Judge Answer Generation Prompt}]
\#\#\# Task Instructions \\
You are a professional and rigorous legal AI assistant. Based on the [Original Query] and the system's [Chain-of-Thought Analysis] provided below, please extract and organize the final answer.

\#\#\# Core Rules: \\
1. Provide a comprehensive summary answer to the original query that is backed by precise evidence, demonstrates clear and profound logical reasoning, reaches a fully correct conclusion, and is articulated with professional phrasing. \\
2. You must eliminate all internal tags and machine retrieval artifacts, such as $<$search$>$, $<$syllogism$>$, and $<$answer$>$. Seamlessly integrate the core substance of relevant legal statutes and analogous cases into your final response. \\
3. Present the final result directly to the user. Do not include machine-perspective pleasantries or boilerplate language like ``According to the system's analysis.''

\#\#\# Input Data \\
- [Original Query]: \textcolor{blue}{\{query\}} \\
- [Chain-of-Thought Analysis]: \textcolor{blue}{\{history\}}

\#\#\# Final Answer:
\end{tcolorbox}
\caption{The prompt template utilized for judge answer generation. Within this template, \textcolor{blue}{\{query\}} denotes the initial legal question inputted by the user, and \textcolor{blue}{\{history\}} represents the comprehensive reasoning chain and multi-turn retrieval trajectory generated by the lawyer agent.}
\label{fig:judge_prompt}
\end{figure*}

\section{Reproducibility and Implementation Details}
\label{app:implementation}
\subsection{Evaluation Configuration}
For all reported evaluations, we set the decoding temperature to $0.0$ and set \texttt{max\_retries} to $5$. For each benchmark setting, the reported score is computed by summing the sample-level scores over the benchmark test set and dividing by the number of evaluated samples.

\subsection{Training Configuration and Reward Design}

\paragraph{Training Configuration.}
Legal-R1 is initialized from Qwen3-8B and trained in two stages. We first perform supervised fine-tuning (SFT) on $D_{\text{SFT}}$ for 3 epochs with a learning rate of $5\times10^{-6}$ and a batch size of 32. We then apply Group Relative Policy Optimization (GRPO) to further optimize the evidence-seeking policy, using a learning rate of $1\times10^{-6}$, a batch size of 32, and 8 rollouts per prompt.

\paragraph{Training Deployment.}
Training is conducted on a single node equipped with four NVIDIA H20 GPUs, each with 141\,GB of VRAM. GPU~0--2 are used for Fully Sharded Data Parallel (FSDP) training and rollout generation. GPU~3 hosts the auxiliary components used during training, including the hybrid retrieval module, which combines BM25 and text2vec and consumes approximately 8\,GB of VRAM. Two additional Qwen3-8B instances are deployed on GPU~3: a Legal Assistant model used for filtering, summarization, and reranking in the retrieval pipeline, and a reward judge model used for reward computation during GRPO. These two models are allocated approximately 40\% and 50\% of GPU~3 memory, respectively.

\paragraph{Reward Components and Weights.}
The trajectory-aware reward described in the Method section combines answer accuracy, trajectory alignment, retrieved-information gain, and query quality with weights $w_1=1.00$, $w_2=0.08$, $w_3=0.32$, and $w_4=0.20$, respectively. The rule-based format term checks valid \texttt{answer}, \texttt{search}, and \texttt{information} tags. The reward further includes a small bonus when the number of searches is close to the reference trajectory, a bounded penalty for overly long trajectories, and the GRPO KL regularization term with $\eta=0.08$.

\subsection{System Deployment and Inference Pipeline}
At inference time, we first initialize the hybrid retriever and launch three model instances with vLLM: (i) the \textit{Legal Assistant Agent} based on Qwen3-8B, which performs filtering, summarization, and reranking over retrieved candidates; (ii) the \textit{Judge Agent} based on Qwen3-8B, which synthesizes the final evidence-grounded answer; and (iii) the \textit{Lawyer Agent}, instantiated by the trained Legal-R1 model, which performs evidence planning and tool invocation.

Given a user question, the Lawyer Agent first analyzes the problem and issues a retrieval request. The hybrid retriever performs coarse-grained retrieval to construct a candidate document pool, after which the Legal Assistant Agent performs fine-grained filtering and reranking. The resulting evidence is returned to the Lawyer Agent, which continues the reasoning--retrieval loop until it determines that sufficient evidence has been collected. The complete retrieval trajectory---including the original question, retrieval actions, intermediate queries, and retrieved evidence---is then passed to the Judge Agent, which synthesizes the collected evidence into the final answer.

\subsection{Serving Cost and Throughput}

For the serving stress test, the complete Legal-R1 inference system is deployed with vLLM on four NVIDIA A100-SXM4-80GB GPUs. Under concurrent inference, the measured GPU memory usage is approximately 72.2\,GB, 70.1\,GB, 70.1\,GB, and 50.7\,GB across the four GPUs, respectively. The end-to-end generation throughput is 239.98 output tokens/s, including internal reasoning and tool-call tokens. These measurements characterize the deployment cost of the complete Legal-R1 pipeline rather than the resource consumption of a single-pass 8B model.

\section{LawBench Efficiency Analysis}
\label{app:efficiency}
LawBench Turns is the average number of retrieval/reasoning rounds during evaluation. LawBench Tokens is the average total token consumption, including input tokens, output tokens, internal reasoning tokens, and retrieval-related tokens. The score column reports the overall average score used in the main comparison, while the two cost metrics are measured on LawBench.

\begin{table*}[t]
\centering
\small
\begin{tabular}{lrrr}
\toprule
\textbf{Model} & \textbf{Avg. Score} & \textbf{LawBench Turns} & \textbf{LawBench Tokens} \\
\midrule
GPT-5.4-Mini & 56.87 & -- & 399.23 \\
AutoRefine-Qwen2.5-7B-Instruct & 56.03 & 4.35 & 37226.08 \\
Qwen3-8B w/ hybrid & 55.91 & 0.35 & 5196.62 \\
Legal-R1 & 60.28 & 2.33 & 10861.85 \\
\bottomrule
\end{tabular}
\caption{Accuracy--cost comparison with LawBench efficiency metrics.}
\label{tab:lawbench_efficiency_added}
\end{table*}

Table~\ref{tab:lawbench_efficiency_added} reports Legal-R1 uses more retrieval/reasoning rounds and tokens than lightweight or shallow-retrieval baselines. For example, relative to Qwen3-8B w/ hybrid, it improves the average score from 55.91 to 60.28 while increasing LawBench Turns from 0.35 to 2.33 and LawBench Tokens from 5196.62 to 10861.85. Compared with the multi-turn AutoRefine-Qwen2.5-7B-Instruct baseline, Legal-R1 attains a higher average score with fewer turns and lower token consumption. Thus, the results characterize a favorable accuracy--cost trade-off among the evaluated multi-turn retrieval systems, rather than eliminating the additional runtime cost of Legal-R1.


\section{LawBench 1-1 Statute Retrieval Case Study}
\label{app:statute_case_study}
We further analyze LawBench 1-1, a subset of 500 questions requiring the original text of legal provisions. After introducing the RAG and reranking pipeline, 364 of the 500 questions receive higher scores. This aggregate result indicates that retrieval augmentation is especially useful when the task requires faithful reproduction of statutory text.

For question ID 101, which asks for the content of Article 14 of the Consumer Rights Protection Law, Legal-R1 retrieves and reproduces the current version of the provision, including the clause that consumers' personal information shall be protected according to law. In contrast, the post-trained model without RAG produces an older version that omits this clause. For question ID 12, which asks for Article 36 of the Law on Sole Proprietorship Enterprises, Legal-R1 answers correctly with retrieval support, whereas the post-trained model without RAG generates the content of Article 32. These examples illustrate that parametric legal knowledge may be outdated or may cite an incorrect provision, while retrieval grounds the answer in current and verifiable legal text.

\section{Full Trajectory on Case Study }
\label{sec:case_study}
\subsection{Case Study}
To intuitively illustrate the autonomous multi-turn reasoning and retrieval capabilities of Legal-R1, we conduct a qualitative case study on a representative real-world legal consultation. As analyzing in Figure~\ref{fig:case_study}, given a user query regarding enforcement penalties, the model does not rely on static memories but independently initiates a sequence of precise retrieval loops to search for statutory criteria and validation procedures for dishonest judgment debtors. Notably, it strategically executes cross-references between general rules (Article 1) and exception clauses (Article 3), while invoking a legal precedent to ground its contextual understanding in judicial practice. Ultimately, through this systematic deliberation, Legal-R1 synthesizes an accurate and legally sound judgment, demonstrating its robust reliability in professional legal applications.

\begin{figure*}[t]
\begin{tcolorbox}[title={Case Study: Autonomous Multi-Turn Retrieval and Reasoning Trajectory}]
\textbf{[User Query]} \\
My friend borrowed 164,000 RMB this January, failed to repay, and was sued. We won the lawsuit, and the court ordered repayment within 25 days. If they still do not pay, will they automatically be blacklisted as a dishonest judgment debtor?

\vspace{0.5em}
\textbf{[Autonomous Reasoning \& Retrieval Trajectory]} \\
\textit{Thinking Step 1:} To address whether non-repayment automatically leads to being blacklisted, I must first locate the precise statutory criteria and procedures regarding the designation of a "dishonest judgment debtor." \\
$\hookrightarrow$ \textbf{Retrieval 1 (Statute):} \textit{Several Provisions of the Supreme People's Court on Publishing Information of the List of Dishonest Judgment Debtors, Article 1.} \\
\textit{Thinking Step 2:} According to Article 1, "having the capability to fulfill but refusing to fulfill" is a core mandatory prerequisite. I need to retrieve further regulatory interpretations on how the court determines "the capability to fulfill." \\
$\hookrightarrow$ \textbf{Retrieval 2 (Statute):} \textit{Several Provisions of the Supreme People's Court on Publishing Information of the List of Dishonest Judgment Debtors, Article 2.} \\
\textit{Thinking Step 3:} I must now locate procedural guidelines or exemption clauses, particularly regarding situations where execution remains incomplete after the court-ordered deadline. \\
$\hookrightarrow$ \textbf{Retrieval 3 (Statute):} \textit{Several Provisions of the Supreme People's Court on Publishing Information of the List of Dishonest Judgment Debtors, Article 3.} \\
\textit{Thinking Step 4:} To validate the interactive boundaries of "non-fulfillment" and "blacklisting" in real judicial practice, I will invoke a precedent retrieval to observe how local courts adjudicate analogous scenarios. \\
$\hookrightarrow$ \textbf{Retrieval 4 (Precedent Case):} \textit{The judgment debtor Xiang Mo failed to fulfill the repayment obligation after the court judgment took effect, and still willfully refused execution after receiving the enforcement notice...}

\vspace{0.5em}
\textbf{[Final Response Generated by Judge Agent]} \\
\textbf{Answer}: No, failure to repay within 25 days does not inevitably lead to being blacklisted. The critical criterion hinges on whether the debtor "has the capability to fulfill but willfully refuses to do so." If the debtor genuinely lacks the capacity to repay (e.g., assets are already sealed by other creditors, or income is insufficient for basic living costs), they may not be blacklisted. Furthermore, according to Article 3 of the \textit{Provisions}, if the debtor provides sufficient collateral, or if their seized/frozen assets are already adequate to liquidate the debt, they shall not be blacklisted. Therefore, the court evaluates this on a case-by-case basis. \\
\textbf{Legal Basis}: Articles 1 and 3 of the \textit{Several Provisions of the Supreme People's Court on Publishing Information of the List of Dishonest Judgment Debtors}.
\end{tcolorbox}
\caption{An idealized snapshot of Legal-R1's autonomous thought-retrieval loop and final response synthesis.}
\label{fig:case_study}
\end{figure*}

\section{Legal Knowledge Base Annotation Details}
\label{sec:appendix_kb}

\subsection{Authority Tier}
We use the level of the adjudicating court as a proxy for the authority of a precedent case. Since judgments from higher-level courts generally carry stronger reference value, each case is mapped to one of three authority tiers and receives a tier-specific score for precedent reranking:
\begin{itemize}
    \item \textbf{Level 1} consists of cases adjudicated by high people's courts. These cases receive an authority score of 1.1 and include 195 cases in LeCaRDv2.
    \item \textbf{Level 2} consists of cases adjudicated by intermediate people's courts. These cases receive an authority score of 1.05 and include 3,731 cases in LeCaRDv2.
    \item \textbf{Level 3} consists of cases adjudicated by primary people's courts or other local tribunals. These cases receive an authority score of 1.0 and include 49,351 cases in LeCaRDv2.
\end{itemize}
These scores provide a lightweight prior rather than a hard filter. Cases from lower courts can still be selected when they are more relevant or contribute distinct sentencing evidence.

\subsection{Punishment Severity Index}
To systematically evaluate judicial outcomes, we develop the Punishment Severity Index (PSI), a computational framework that converts heterogeneous sentencing descriptions into a continuous numerical scale.This ensures that linguistic variations in judicial texts (e.g., ``three years'' vs. ``36 months'') are standardized into uniform computational inputs.

The core of the PSI(Punishment Severity Index) is based on a weighted temporal-equivalent model. Main penalties are first converted into a day-equivalent baseline ($D$), which is then adjusted by a Severity Coefficient ($\alpha$) and a Probation Adjustment Factor ($\phi$). Specifically, fixed-term imprisonment serves as the baseline ($\alpha=1.0$), while criminal detention and public surveillance are assigned lower weights ($\alpha=0.8$ and $0.3$, respectively) to reflect their lower degree of liberty deprivation. Furthermore, if a suspended sentence is detected, a multiplier of $\phi=0.2$ is applied. For extreme non-temporal penalties, such as life imprisonment or the death penalty, the index assigns discrete, high-order ordinal values (e.g., $15,000$ to $100,000$) to maintain the hierarchical integrity of the penal system.

Finally, the index incorporates Supplementary Penalties through additive weighting. To account for the diminishing marginal impact of monetary sanctions, fines and property forfeitures are integrated using a logarithmic scaling function: $S_{fine} = 20 \times \log_{10}(\text{amount} + 1)$. Other sanctions, including the deprivation of political rights and deportation, are assigned fixed weights based on their statutory duration or severity. The total PSI is the cumulative sum of these components:$$PSI = (D \cdot \alpha \cdot \phi) + \sum S_{supplementary}$$This integrated metric provides a robust basis for subsequent statistical analysis and cross-case comparisons of sentencing severity.

\begin{table*}[t] 
\centering
\begin{CJK*}{UTF8}{gbsn}
\scalebox{0.8}{
\begin{tabular}{p{7.5cm} l c}
\toprule
\textbf{Penalty Case} & \textbf{Calculation Process} & \textbf{PSI Result} \\
\midrule
死刑 \newline (Death penalty) & $100,000$ & 100,000.0 \\
无期徒刑，剥夺政治权利终身 \newline (Life imprisonment with lifelong deprivation of political rights) & $15,000 + 2,000$ & 17,000.0 \\
\midrule
有期徒刑十五年 \newline (Fixed-term imprisonment of fifteen years) & $(15 \times 12 \times 30) \times 1.0$ & 5,400.0 \\
有期徒刑三年，缓刑四年 \newline (Fixed-term imprisonment of three years with a four-year suspension of sentence) & $(3 \times 12 \times 30) \times 1.0 \times 0.2$ & 216.0 \\
\midrule
拘役六个月，并处罚金一万元 \newline (Criminal detention of six months, with a concurrent fine of 10,000 RMB) & $(6 \times 30) \times 0.8 + 20 \times \log_{10}(10,000 + 1)$ & 224.0 \\
管制一年 \newline (Public surveillance for one year) & $(1 \times 12 \times 30) \times 0.3$ & 108.0 \\
\midrule
驱逐出境 \newline (Deportation) & $1,000$ & 1,000.0 \\
免予刑事处罚 \newline (Exemption from criminal punishment) & $0$ & 0.0 \\
\bottomrule
\end{tabular}}
\caption{Examples of PSI Calculation for Different Penalties.} \label{tab:psi_cases}
\end{CJK*} 
\end{table*}

\section{Details of Training Data Construction}
\label{app:data_construction}

This appendix provides additional details on the construction of the post-training dataset used in our legal agent model. The goal of data construction is to obtain high-quality, evidence-grounded, and tool-use-oriented legal QA instances that can improve the model's ability to understand legal facts, invoke external tools, retrieve relevant legal knowledge, and generate reliable answers.

\subsection{Data Sources and Task Coverage}
\begin{table}[ht]
\centering
\scalebox{0.8}{
\renewcommand\tabcolsep{2.8pt}
\begin{tabular}{llr}
\toprule
\textbf{Main Category} & \textbf{Sub-category} & \textbf{Count} \\
\midrule
\multirow{3}{*}{Knowledge Retrieval} & Clause Recall & 500 \\
 & Clause Retrieval & 800 \\
\cmidrule{2-3}
 & \textit{Subtotal} & \textit{1,300} \\
\midrule
\multirow{12}{*}{Understanding} & Case Similarity & 101 \\
 & Dispute Identification & 300 \\
 & Concept Comprehension & 1,000 \\
 & Marriage Assessment & 454 \\
 & Subject Identification & 1,000 \\
 & Reading Comprehension & 1,000 \\
 & Event Detection & 1,000 \\
 & Trigger Extraction & 1,000 \\
 & Relation Understanding & 68 \\
 & Plaintiff Opinion & 1,000 \\
 & Document Correction & 500 \\
\cmidrule{2-3}
 & \textit{Subtotal} & \textit{7,423} \\
\midrule
\multirow{8}{*}{Reasoning} & Charge Prediction & 460 \\
 & Charge with law Prediction & 426 \\
 & Amount Calculation & 401 \\
 & Sentence Prediction & 420 \\
 & Scenario Analysis & 1,000 \\
 & Legal Calculation & 1,000 \\
 & Multi-hop Q\&A & 1,000 \\
\cmidrule{2-3}
 & \textit{Subtotal} & \textit{4,707} \\
\midrule
\multirow{6}{*}{Generation} & Document Summarization & 1,000 \\
 & Indictment Generation & 420 \\
 & Contract Generation & 8 \\
 & Opinion Summarization & 1,000 \\
 & Term Translation & 500 \\
\cmidrule{2-3}
 & \textit{Subtotal} & \textit{2,928} \\
\midrule
\multirow{4}{*}{Consultation} & User Consultation & 2,000 \\
 & Evidence Opinion & 158 \\
 & Case Judgment & 424 \\
\cmidrule{2-3}
 & \textit{Subtotal} & \textit{2,582} \\
\midrule
\textbf{Total} & & \textbf{18,940} \\
\bottomrule
\end{tabular}}
\caption{Distribution of Dataset Categories and Sub-categories.}
\label{tab:dataset_stats_v2}
\end{table}

Our post-training dataset consists of 18,940 structured QA instances, covering 5 major legal domains and 29 fine-grained subcategories, detailed in Table~\ref{tab:dataset_stats_v2}. The raw materials are collected from a combination of public legal NLP benchmarks, authoritative legal examination resources, and domain-specific synthetic problem sets. Specifically, the primary data sources include the Chinese AI and Law Challenge (CAIL), DISC-LawLLM, CrimeKgAssitant, and the National Unified Legal Professional Qualification Examination. These resources provide diverse legal materials, including statutory provisions, case facts, legal reasoning questions, criminal law problems, civil law problems, procedural law questions, and examination-style multiple-choice questions.

To improve task diversity, we organize the dataset according to a task-oriented configuration profile. Each task type is associated with a dedicated task description, input format, output format, and generation prompt. This design allows the pipeline to convert heterogeneous source materials into structured training instances under a unified schema. The resulting QA pairs cover both open-ended verifiable questions and legal selection tasks, enabling the model to learn different forms of legal reasoning, including fact understanding, legal norm matching, punishment prediction, legal element identification, and multi-step evidence-grounded answering.

\subsection{Overview of the Automated Construction Pipeline}

{{
To construct the dataset at scale while maintaining data quality, we design an end-to-end automated synthesis and purification pipeline. We use DeepSeek-V3 with temperature $0.1$ to generate verifiable QA pairs from legal materials selected by task templates. Given raw legal materials, the pipeline first determines whether each source text is suitable for a target task, then synthesizes candidate QA pairs, and finally applies automatic quality-control procedures. It rejects instances with invalid JSON, missing answers, task--material mismatches, or answer leakage. The overall pipeline consists of task-driven applicability gating, fingerprint-based contamination filtering, automated answer-leakage suppression, and rule-driven distractor design.

}}

Formally, let $\mathcal{D}_{raw}$ denote the collection of raw legal materials, and let $\mathcal{T}$ denote the set of target task types. For each source text $x \in \mathcal{D}_{raw}$ and each task type $\tau \in \mathcal{T}$, the pipeline first evaluates whether $x$ contains sufficient information to support task $\tau$. If the source text passes this applicability check, the model generates one or more candidate QA instances. Each candidate instance is then validated by contamination detection and task-specific quality filters. Only instances that pass all filtering steps are retained in the final post-training dataset $\mathcal{D}_{train}$.

\subsection{Task-Driven Two-Stage Applicability Gating}

A major challenge in legal data synthesis is that not every source text is suitable for every downstream task. For example, a statutory article may be appropriate for legal norm identification but may not contain sufficient factual information for punishment prediction. Similarly, a case fact description may support legal issue analysis but may not provide enough explicit evidence for a multiple-choice question about procedural requirements. Directly generating QA pairs from mismatched source materials can therefore introduce unanswerable questions, hallucinated answers, or weakly grounded supervision signals.

To address this issue, we implement a task-driven two-stage applicability gating mechanism before QA synthesis. In the first stage, the pipeline performs a coarse task-material matching check based on metadata, source type, and task description. This step filters out clearly incompatible combinations, such as applying a sentencing-oriented task to a source text that contains only abstract legal definitions.

In the second stage, the pipeline performs a model-based feasibility pre-vetting. Given a source text and the target task description, the model is instructed to explicitly judge whether the material contains sufficient factual predicates, legal entities, or normative conditions required by the task. The model must output a binary decision together with a brief justification. If the material lacks key factual disclosures, contains insufficient legal context, or cannot support a verifiable answer, the corresponding task is skipped and recorded as a discarded instance.

This two-stage design filters low-quality candidates before generation rather than after generation. It reduces unnecessary synthesis costs and prevents the pipeline from producing superficially fluent but legally unsupported QA pairs.

\subsection{Fingerprint-Based Data Contamination Filtering}

To reduce the risk of benchmark leakage, we incorporate a fingerprint-based contamination filtering module. Before running the data synthesis pipeline, all designated evaluation benchmarks are cleaned and normalized to construct a global fingerprint library. The normalization process removes redundant whitespace, standardizes punctuation, and unifies common formatting variants. 
{{
Each benchmark sample is decomposed into exact 20-gram fingerprints.
}}

During data construction, every candidate source text is compared against the fingerprint library before it is used for QA synthesis. If the source text contains an 20-gram segment that overlaps with the held-out benchmark fingerprints above a predefined threshold, the sample is treated as potentially contaminated and is removed from the generation process. This step is applied at the source-text level rather than only at the final QA-pair level, because a generated question may paraphrase a contaminated source and become difficult to detect after synthesis.

The contamination filter is designed to be conservative. Its purpose is not to remove all topically similar legal materials, since different legal cases may naturally involve similar statutes or legal concepts. Instead, it aims to eliminate near-duplicate or directly overlapping textual evidence from held-out benchmarks, thereby reducing the possibility that evaluation performance is inflated by memorization.

\subsection{Automated Syntax Parsing and Answer Leakage Suppression}

For open-ended verifiable QA tasks, it is important that the question does not reveal the ground-truth answer. This is especially critical in agentic retrieval settings, where the model is expected to decide whether to invoke tools, retrieve legal evidence, and reason from the retrieved information. If the answer is already exposed in the question, the instance becomes trivial and no longer provides useful supervision for retrieval or legal reasoning.

We therefore design an automated answer-leakage suppression module. Given a generated QA pair $(q, a)$, the module performs two layers of validation. First, it checks whether the answer string $a$ appears directly in the question $q$. This direct substring matching is effective for detecting explicit leakage, such as questions that already contain the target charge, statute number, sentencing result, or legal conclusion.

Second, the module applies regular-expression-based pattern matching to identify sensitive answer-bearing values. For example, in punishment-related questions, the parser detects sentencing expressions, penalty types, durations, fines, suspension conditions, and other legally meaningful values. If such values appear in the question and match the expected answer format, the instance is flagged as answer-leaking and discarded. This rule is particularly useful when the answer is not copied verbatim but appears in a slightly different surface form.

After leakage filtering, the remaining questions are more suitable for evaluating and training the model's ability to infer answers from legal evidence rather than relying on superficial cues in the question itself.

\subsection{Rule-Driven Heuristic Distractor Design for Legal Selection Tasks}

Multiple-choice legal questions require high-quality distractors. If distractor options are too obviously wrong, the task becomes overly simple; if they are semantically overlapping with the correct answer, the question may become ambiguous or legally invalid. To avoid these issues, we do not rely solely on unconstrained model generation for option construction. Instead, we design a rule-driven heuristic distractor generation module.

Given a source text and the corresponding correct answer, the module constructs negative options through two main corruption strategies.

\paragraph{Value Swapping.}
The first strategy creates distractors by swapping authentic values across legal entities or factual roles. For example, if a case description contains multiple actors, amounts, dates, or sentencing results, the algorithm reassigns a correct value from one entity to another incorrect entity. Since the values are drawn from the original text, the generated distractors remain plausible at the surface level. However, because the entity-value alignment is corrupted, the distractors are legally incorrect.

\paragraph{Wrong Label Assignment.}
The second strategy constructs distractors by assigning correct textual entities to incorrect legal labels. For example, a factual element that corresponds to one legal role, charge, or statutory condition may be deliberately mapped to another label. This creates options that are lexically grounded in the source material but invalid under legal logic.

After generating candidate distractors, the module applies validity checks to ensure that each distractor is distinct from the correct answer and from other options. The correct option and the distractors are then randomly shuffled and mapped to an answer index, such as A, B, C, or D. This randomization prevents the model from exploiting positional bias during training.

\subsection{Quality Control and Final Dataset Assembly}

After the above filtering and validation steps, all retained QA instances are converted into a unified training format. Each instance contains the task instruction, question, answer, optional choices, supporting evidence, and task metadata. For tool-use-oriented tasks, the instance may additionally include tool invocation annotations, intermediate evidence, or retrieval-oriented supervision signals.

The final dataset assembly process follows three principles. First, each instance should be answerable from its associated evidence or legal knowledge source. Second, the question should not contain direct answer leakage or shortcut cues. Third, the answer should be legally valid, concise, and consistent with the intended task type. Instances that violate any of these principles are removed.

Through this automated construction pipeline, we obtain a post-training dataset that is diverse in legal domain coverage, structured in task format, and reliable in supervision quality. The resulting dataset supports the model's post-training for legal question answering, legal retrieval, tool invocation, and evidence-grounded reasoning.

\section{The Use of Large Language Models}
\paragraph{Using an LLM to help with paper writing}
During the preparation of this work, the authors utilized Large Language Models for language polishing, improving the structural clarity of the manuscript, and refining the formal expression of individual sentences. The use of Large Language Models did not influence the substantive content of the study and served solely as a writing aid.

\section{Potential Risks}
While Legal-R1 significantly enhances the accuracy of legal question answering through its active evidence-seeking framework, we acknowledge several potential risks associated with its deployment:

\begin{itemize}
    \item \textbf{Risk of Hallucination and Misguidance:} As an LLM-based system, Legal-R1 may still suffer from subtle hallucinations or generate plausible-sounding but incorrect legal reasoning in highly complex or edge cases. If users, especially non-professionals, rely entirely on the model's outputs for real-world litigation or legal decisions without professional verification, it could lead to severe legal or financial consequences.
    \item \textbf{Over-Reliance on the System:} The structured nature and highly confident tone of the multi-turn trajectories generated by Legal-R1 might inadvertently lead users to treat it as an absolute legal authority, reducing human legal practitioners' critical oversight.
    \item \textbf{Jurisdictional and Temporal Limitations:} The current model and its underlying heterogeneous knowledge base are strictly tailored to the contemporary Chinese legal framework. Applying this model to foreign jurisdictions, alternative legal systems (e.g., common law), or outdated legal contexts could produce entirely invalid results.
\end{itemize}

To mitigate these risks, we explicitly emphasize that Legal-R1 is designed strictly as a \textit{supportive tool} to assist legal professionals in research and evidence collection, rather than a replacement for qualified legal counsel. We recommend that any downstream deployment of this system incorporate clear user disclaimers and human-in-the-loop validation mechanisms to ensure safety and compliance.

\end{document}